%% file: main.tex
\PassOptionsToPackage{table,dvipsnames}{xcolor}
\documentclass{article}

\PassOptionsToPackage{authoryear,square}{natbib}

\usepackage[preprint]{arxiv}

\usepackage[utf8]{inputenc} %
\usepackage[T1]{fontenc}    %
\usepackage{hyperref}       %
\usepackage{url}            %
\usepackage{booktabs}       %
\usepackage{amsfonts}       %
\usepackage{nicefrac}       %
\usepackage{microtype}      %
\usepackage{array}
\usepackage{xcolor}         %
\usepackage{amsmath}
\usepackage{multirow}  %
\usepackage{adjustbox}
\usepackage{makecell}
\usepackage{subcaption}
\usepackage{xspace}
\usepackage{amssymb}
\usepackage{adjustbox}
\usepackage{mathtools}
\usepackage{pifont}
\usepackage{tikz}
\usepackage{graphicx}
\usepackage{siunitx}
\usepackage{tcolorbox}
\usepackage{caption}
\usepackage{titletoc}
\usepackage{placeins}

\definecolor{wongOrange}{HTML}{E69F00}
\definecolor{wongSky}{HTML}{56B4E9}
\definecolor{wongBlue}{HTML}{0072B2}
\definecolor{wongGrey}{HTML}{999999}
\definecolor{tabgray}{gray}{0.4}
\definecolor{selected}{HTML}{E7FFE7}
\colorlet{ontarget}{wongSky!30!white}
\colorlet{defaultcolor}{wongOrange!70!black}

\newcommand\plainthanks[1]{%
\begin{NoHyper}\thanks{#1}\end{NoHyper}%
}

\newcolumntype{C}{>{\centering\arraybackslash}c}
\newcolumntype{N}{>{\centering\arraybackslash}m{1.05cm}}
\newcolumntype{G}{>{\color{tabgray}\centering\arraybackslash}m{1.05cm}}

\newcommand{\rank}[1]{{\color{gray}\footnotesize\textsuperscript{#1}}}

\newcommand{\coremetric}{\textsc{Core}\xspace}
\newcommand{\coremetricavg}{\textsc{Core} (Avg)\xspace}
\newcommand{\extendedmetric}{\textsc{Noncore}\xspace}

\newcommand{\bydefault}[1]{\textcolor{defaultcolor}{\textbf{In practice:}} #1}

\newcommand{\keybox}[1]{%
  \begin{tcolorbox}[
    colback=orange!8,
    colframe=orange!8,  %
    boxrule=0pt,
    left=3pt,
    right=3pt,
    top=3pt,
    bottom=3pt,
    arc=0pt,
    boxsep=3pt,
  ]
  \textbf{Takeaway:} #1
  \end{tcolorbox}
}

\newcommand{\highlight}[2][yellow]{%
  {\setlength{\fboxsep}{1.5pt}\colorbox{#1}{#2}}%
}

\title{Language Models Improve When \\ Pretraining Data Matches Target Tasks}

\author{%
\textbf{David Mizrahi}$^{1}$ \qquad \textbf{Anders Boesen Lindbo Larsen}$^{1}$ \qquad \textbf{Jesse Allardice}$^{1}$  \vspace{0.5em} \\
\setcounter{footnote}{1}  %
\textbf{Suzie Petryk}$^{1}$ \qquad \textbf{Yuri Gorokhov}$^{1}$ \qquad \textbf{Jeffrey Li}$^{2}$\plainthanks{Work done during an internship at Apple.} \qquad \textbf{Alex Fang}$^{1, 3}$\footnotemark[2]   \vspace{0.5em} \\
\textbf{Josh Gardner}\plainthanks{Work done while at Apple. Josh Gardner is now at Anthropic.} \qquad \textbf{Tom Gunter}\footnotemark[3] \qquad  \textbf{Afshin Dehghan}$^{1}$ \vspace{0.5em} \\
$^1$Apple \qquad $^2$University of Washington \qquad $^3$Stanford  \\ 
}

\begin{document}

\maketitle
\enlargethispage{2\baselineskip}

\begin{abstract}

\input{sec/0_abstract}

\end{abstract}

\setcounter{footnote}{0}

\vspace{-0.2\baselineskip}
\input{figures/main/pull_fig}

\section{Introduction}
\input{sec/1_introduction}

\section{Related work}
\input{sec/2_related_work}

\section{Method}
\input{sec/3_method}

\section{Experimental setup}
\input{sec/4_experimental_setup}

\section{Results}
\label{sec:results}
\input{sec/5_results}

\section{Insights from scaling laws}
\label{sec:scaling_insights}
\input{sec/6_scaling_law_insights}

\section{Discussion \& Conclusion}

\input{sec/7_discussion_conclusion}

\medskip

{
\small
\bibliographystyle{plainnat}
\bibliography{references}
}
\newpage
\appendix
\input{sec/8_appendix}

\end{document}

%% file: sec/0_abstract.tex
\noindent Every data selection method inherently has a target. In practice, these targets often emerge implicitly through benchmark-driven iteration: researchers develop selection strategies, train models, measure benchmark performance, then refine accordingly. This raises a natural question: what happens when we make this optimization explicit? To explore this, we propose benchmark-targeted ranking (BETR), a simple method that selects pretraining documents based on similarity to benchmark training examples. BETR embeds benchmark examples and a sample of pretraining documents in a shared space, scores this sample by similarity to benchmarks, then trains a lightweight classifier to predict these scores for the full corpus. \newline
We compare data selection methods by training over 500 models spanning $10^{19}$ to $10^{22}$ FLOPs and fitting scaling laws to them. From this, we find that simply aligning pretraining data to evaluation benchmarks using BETR achieves a 2.1x compute multiplier over DCLM-Baseline (4.7x over unfiltered data) and improves performance on 9 out of 10 tasks across all scales. BETR also generalizes well: when targeting a diverse set of benchmarks disjoint from our evaluation suite, it still matches or outperforms baselines. Our scaling analysis further reveals a clear trend: larger models require less aggressive filtering.  Overall, our findings show that directly matching pretraining data to target tasks precisely shapes model capabilities and highlight that optimal selection strategies must adapt to model scale.

%% file: figures/main/pull_fig.tex
\begin{figure}[!htb]
  \centering
  \includegraphics[width=\columnwidth]{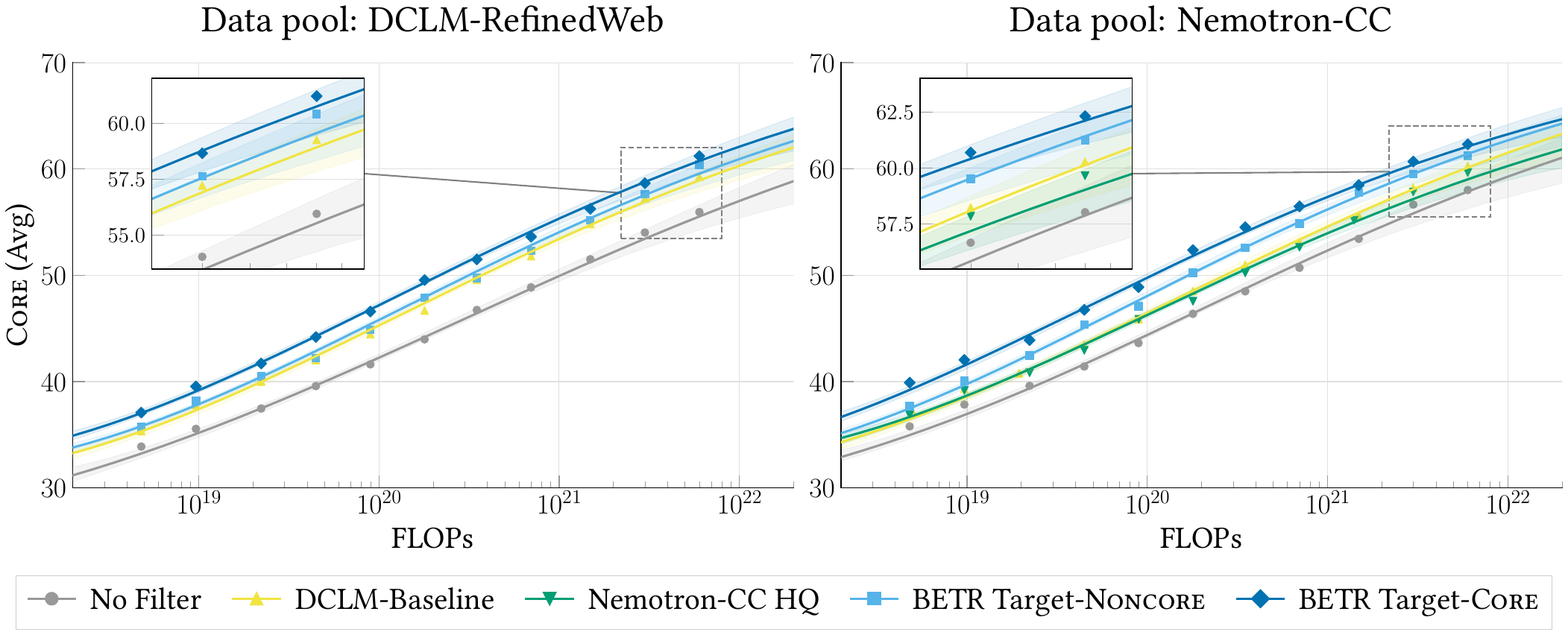}
  \caption{\textbf{Benchmark-targeted ranking (BETR) achieves a 1.8x--2.8x compute multiplier over strong baselines.} Scaling curves show accuracy on \coremetric (10 standard benchmarks) at compute-optimality from $10^{19}$ to $10^{22}$ FLOPs. \textcolor{wongBlue}{Target-\coremetric} directly optimizes for evaluated benchmarks, while \textcolor{wongSky}{Target-\extendedmetric} targets distinct benchmarks. Both outperform DCLM-Baseline at all scales.}
  \label{fig:pull_figure}
\end{figure}

%% file: sec/1_introduction.tex
Pretraining data fundamentally shapes the capabilities of large language models (LLMs). Data improvements have been a key driver in LLM advances~\citep{openai2024gpt4o,anthropic2025claude4,geminiteam2025gemini25,deepseek2024v3,meta2024llama3,qwen2025v3}, with recent document-level data selection methods~\citep{penedo2024fineweb, li2024dclm, su2024nemotron} showing that carefully selected datasets can significantly improve performance across a wide range of tasks. However, current methods largely rely on researcher intuition about what constitutes quality, whether using LLMs to make quality judgments based on implicit criteria~\citep{penedo2024fineweb,kong2024large} or manually selecting examples of ``high-quality'' documents for training classifiers, then iterating based on benchmark results~\citep{li2024dclm}. This makes datasets difficult to systematically improve, as the very notion of ``quality'' remains loosely defined.

This ambiguity arises because \textbf{quality is inherently task-dependent}. Documents valuable for one task often differ from those useful for another. Without specific objectives, ranking diverse documents becomes meaningless: how would one compare the relative quality of a mathematical proof, a poem, and a news article? In practice, however, data selection methods already have implicit objectives: they consistently iterate based on performance measured against standard benchmarks~\citep{li2024dclm}. This raises a natural question: if benchmark performance implicitly guides the development of data selection methods, what happens when we \textbf{explicitly align} pretraining data with these benchmarks?

To explore this, we propose \textbf{benchmark-targeted ranking (BETR)}, which directly aligns pretraining data with benchmarks by ranking documents based on their similarity to benchmark examples. BETR operates in three steps: (1) embedding benchmark examples together with a small sample of pretraining documents, (2) scoring documents based on their similarity to benchmark examples, and (3) training an element-wise document scorer to efficiently predict these scores for the full data pool. To ensure evaluation integrity, we target only the training sets of benchmarks and perform careful decontamination against test sets.

We evaluate BETR by training over 500 models spanning compute budgets from $10^{19}$ to $10^{22}$ FLOPs, and find that explicitly aligning pretraining data with benchmarks yields consistent improvements, achieving \textbf{1.8--2.8x compute multipliers} over state-of-the-art baselines across all scales tested. These results demonstrate that direct data-task matching provides better performance at fixed compute budgets, or equivalently, matches baseline performance using only 35--55\% of the compute.

Beyond performance improvements, BETR shows how \textbf{benchmark selection fundamentally shapes model capabilities}. The method can create specialist models by targeting individual benchmarks, or generalist models by targeting diverse benchmark sets---each with distinct but complementary strengths. When trained toward standard evaluation benchmarks, models consistently achieve top performance on those tasks, but at the cost of broader capabilities. In contrast, when trained toward an entirely different set of 39 diverse benchmarks, models remain competitive on our evaluation tasks despite never explicitly optimizing for them. This suggests that diverse benchmark targeting fosters broadly useful capabilities rather than narrow, task-specific behaviors.

Our scaling analysis moves beyond the single-scale evaluations common in previous work and reveals a clear trend: \textbf{optimal data filtering strategies vary predictably with model scale}. Specifically, smaller models perform best with aggressively filtered data (top 3\% at $10^{20}$ FLOPs), while larger models benefit from greater data diversity (top 30\% at $10^{23}$ FLOPs). The consistency of this relationship across data pools suggests that fixed filtering rates currently used in practice may be suboptimal when applied across varying compute budgets.

While the premise that benchmark-relevant data improves benchmark performance may seem self-evident, its implications (and effective execution) are not. Our work highlights an often-overlooked reality: benchmarks do not merely measure progress, they implicitly guide it. By explicitly aligning data with benchmarks, BETR provides a practical means to improve model training and allows us to understand how benchmark choices shape---and constrain---progress in language modeling.

%% file: sec/2_related_work.tex
\subsection{Data selection for language models}

Web crawl data undergoes multiple preprocessing stages, including HTML parsing, text cleaning, deduplication, and coarse filtering, with the goal to select the parts of the Web that are relevant for pretraining while filtering to an appropriate size for the given compute budget. For a broad coverage of such pipelines, we refer to Gopher \citep{rae2021gopher}, DCLM \citep{li2024dclm}, FineWeb \citep{penedo2024fineweb}, and the survey by \citet{albalak2024survey}. The data selection methods we discuss operate at the end of this pipeline.

These data selection methods can be organized along two dimensions:\textbf{ (1) the granularity at which they operate}, from group-level to document-level selection, and \textbf{(2) how directly they optimize for target tasks}. In practice, direct optimization becomes less tractable at finer granularities, requiring different strategies at each level.

\textbf{Group-level selection} methods work with grouped data, where groups may be defined by dataset boundaries~\citep{gao2020pile}, topics and formats~\citep{wettig2025organize}, web domains~\citep{thrush2024improving}, or learned clusters~\citep{diao2025climb}. These methods aim to find group-level mixture weights that maximize performance on target tasks. The approaches range from manual mixtures based on relative size and expected importance~\citep{brown2020gpt3} to learned methods. Learned approaches can be either offline, where mixtures are tuned through separate experiments~\citep{xie2023doremi,fan2023doge,liu2024regmix,grangier2024task,diao2025climb,shukor2025scalinglawsoptimaldata}, or online, with mixture weights dynamically adjusted during model training~\citep{chen2023skill,albalak2023efficient,jiang2024adaptive,chen2024aioli}. Despite their sophisticated optimization procedures, these methods remain constrained, as they treat all documents within a group as equivalent, and the groupings themselves (whether by dataset, domain, format, or topic) are largely heuristic.

\textbf{Document-level selection} methods address the limitations of group-level approaches by scoring at the level of individual documents, but face a different challenge: with billions of documents in web-scale corpora, direct optimization for target tasks becomes computationally intractable. Instead, these methods typically apply element-wise scoring models (e.g. \citep{fang2023data}) that can efficiently label each document. The scoring approaches vary widely: optimization algorithms that learn scorers for specific objectives~\citep{yu2024mates,chen2024take,brandfonbrener2024color,shum2025predictive,gu2025data, calian2025datarater}, perplexity-based filtering~\citep{marion2023less,ankner2024perplexed} (which has been extended to token-level selection~\citep{lin2024rho}), domain matching to either ``high-quality'' web sources~\citep{xie2023data} or specific domains~\citep{shao2024deepseekmath,zhu2024deepseekcoderv2}, LLM-based quality scoring~\citep{sachdeva2024train,wettig2024qurating,penedo2024fineweb,kong2024large}, classifiers trained to distinguish manually curated ``high-quality'' examples from general web ~\citep{li2024dclm}, and ensemble methods that combine multiple scores and annotations~\citep{su2024nemotron,hojel2025essential}. While the more sophisticated approaches seem appealing, DCLM showed that a simple FastText classifier~\citep{joulin2016fasttext} outperforms embedding-based methods~\citep{abbas2023semdedup,xiao2024cpack}, perplexity filtering~\citep{wenzek2019ccnet}, and LLM scoring~\citep{sachdeva2024train}. However, finding the right positive examples (Reddit ELI5~\citep{fan2019eli5} + OpenHermes-2.5~\citep{openhermes}) involved iterative experimentation where manually chosen candidates were systematically tested against benchmarks~\citep{li2024dclm}.

Both group- and document-level methods ultimately optimize for benchmark performance, just through different levels of indirection.\footnote{For an informal discussion on this point, see \citet{beyer2025tweet}.} Group-level methods optimize mixture weights directly but only at coarse granularity, while document-level methods usually rely on proxies refined through benchmark feedback. BETR explores a more direct path by explicitly aligning pretraining documents with benchmark training examples. This allows for document-level selection without the need for iterative refinement or assumptions about what constitutes quality.

\subsection{Data scaling laws}
\textbf{Scaling laws} enable prediction of model behavior across compute scales, initially focused on how loss decreases with model size and training tokens~\citep{kaplan2020scaling,hoffmann2022training}.  Recent work shows that different datasets exhibit distinct scaling properties: compute-optimality varies by data source~\citep{pandey2024gzip,bi2024deepseek}, and losses from different datasets transfer with predictable relationships~\citep{brandfonbrener2024loss,mayilvahanan2025llms}. Special attention has been given to data-constrained regimes where limited data must be repeated, both for language models~\citep{muennighoff2023scaling} and contrastive vision-language models~\citep{goyal2024scaling}. For vision-language models, \citet{goyal2024scaling} found that larger CLIP models benefit from less aggressive filtering when data is constrained. We find that optimal filtering rates for BETR change predictably with scale, extending \citet{goyal2024scaling}'s observations to language model pretraining and showing this pattern appears even before entering data-constrained regimes (i.e., without multiple training epochs).

\textbf{Comparing datasets at scale} presents unique challenges since pretraining loss alone does not always predict downstream performance~\citep{schaeffer2024has}, and because each dataset has its own scaling slope and irreducible loss floor~\citep{pandey2024gzip}. Recent methods address this through two-step prediction: first predicting loss from scale, then mapping loss to benchmark accuracy~\citep{gadre2024language,meta2024llama3,bhagia2024establishing}, which we also employ in this work. While DataDecide~\citep{magnusson2025datadecide} compared single-scale and multi-scale evaluation approaches, ultimately advocating for single-scale evaluation, we employ both: single-scale experiments for rapid comparison and ablations, and scaling laws to verify relative rankings across scales and quantify compute multipliers between methods.

%% file: sec/3_method.tex
\subsection{Motivation and problem setting}

\input{figures/main/method_fig}

What happens if we directly align pretraining data with target tasks? To answer this, we need a principled method for selecting pretraining documents that explicitly leverages benchmarks. Such an approach should scale easily to web-scale corpora (>10B documents) and operate at the document-level, rather than at the topic- or domain-level, given the advantages of document-level selection shown by \citet{li2024dclm, penedo2024fineweb, su2024nemotron}.

With these goals in mind, we frame data selection as a document ranking problem. Ranking documents requires clearly defined targets, since without targets it is unclear what makes one document better or more relevant than another. Benchmarks provide these targets by representing desired tasks and capabilities.
Formally, given a large pool of pretraining documents $\mathcal{D} = \{d_1, \dots, d_N\}$ and a much smaller set of benchmark training examples $\mathcal{B} = \{b_1, \dots, b_M\}$ representing our targets, our goal is to learn a scoring function $s: \mathcal{D} \rightarrow [0,1]$ that measures each document’s contribution to benchmark performance. These scores enable us to rank and filter the training pool, for example by selecting only the highest-scoring documents for model training.

\subsection{Method overview}

Given access to both a pretraining data pool and target benchmark examples, benchmark-targeted ranking (BETR) operates in three steps, illustrated in Figure~\ref{fig:method}. First, we embed both benchmark examples and a sample of pretraining documents in a shared space. Second, we score the sampled documents based on their proximity to benchmark examples. Third, we train a classifier on these scored documents to efficiently predict scores for the entire data pool, and finally use these predicted scores for ranking and filtering.

\subsection{Embedding documents and benchmarks}
\label{sec:embedding_documents}

To rank pretraining documents by similarity to benchmarks, we first embed both into a shared space. This step involves three key design choices: how many pretraining documents to sample, how to represent benchmark targets, and which embedding model to use.

\textbf{Document sampling.} While we ultimately need scores for billions of pretraining documents, computing similarities between every pretraining document and every benchmark example is computationally prohibitive. Instead, we score a representative sample directly and later train a classifier to predict these scores for the full corpus (see  Section \ref{sec:score_extension}). This sample size needs to capture the variety of web text without requiring excessive compute for the initial scoring phase.
\newline \bydefault{We sample 10M documents (less than 0.1\% of the total data pool).}

\textbf{Target granularity.} We must also decide how to aggregate benchmark examples into embedding targets, which controls the granularity of these targets. Options range from embedding each benchmark example individually (most granular), to computing one centroid per benchmark, to using a single global reference point (least granular).
\newline \bydefault{We embed each example individually, as we found it performs best empirically (see Section~\ref{sec:main_ablations}).}

\textbf{Embedding model.} The embedding space must meaningfully capture relationships between web documents and benchmark examples. We use BERT-like transformer encoders~\citep{devlin2019bert}, as they are simple and widely-used for document retrieval. However, our approach is more general: it can use any method capable of ranking documents relative to a given target, such as rerankers or alternative embedding spaces~\citep{thrush2024improving, shum2025predictive}.
\newline \bydefault{We use Arctic-Embed L 2.0~\citep{yu2024arctic} for main results and GTE Large v1.5~\citep{li2023gte,zhang2024mgte} for ablations (see Section~\ref{sec:main_ablations}).}

\subsection{Scoring documents by similarity to benchmarks}

Given embeddings for documents and benchmark examples, the next step is to measure how closely each document aligns with these benchmarks. Since each document is compared against thousands of benchmark examples, it is not immediately clear how to aggregate this information into a single, useful document score. Should we reward documents that are extremely close to just one benchmark example, or those that are moderately close to many?

To address this, we define a scoring function based on similarity ranks. Formally, each benchmark example ranks all documents by similarity, giving each document $d_j$ a rank $r_{ij}$ relative to benchmark example $i$ (i.e., how highly benchmark example $i$ ranks document $j$ compared to other documents). We then apply a function $v(r)$ that assigns a value to each rank. These values are aggregated either by averaging, $S_j = \text{mean}_i\{v(r_{ij})\}$, or by taking the maximum value, $S_j = \max_i\{v(r_{ij})\}$.\footnote{This approach implicitly performs a form of distribution matching. We also explored explicit distribution matching methods (e.g. optimal transport~\citep{peyre2019computational}), but found the rank-based approach simpler and easier to scale.}

The choice of the value function $v(r)$ and aggregation method creates a spectrum of scoring behaviors. Mean aggregation with $v(r) = \log_2(1/r)$ rewards documents with high average relevance across benchmark examples. Max aggregation rewards documents with very high relevance to any single benchmark example. Between these extremes, steeper functions like $v(r) = 1/r$ increasingly favor the best matches.

\bydefault{We use max aggregation,  which scores documents by their best rank across all benchmark examples (see Section~\ref{sec:main_ablations} and Figure~\ref{fig:scoring_distributions}).}

\input{figures/main/rank_sim_distribution}

\subsection{Predicting scores for the full data pool}
\label{sec:score_extension}

As mentioned in Section~\ref{sec:embedding_documents}, directly computing similarities between all pretraining documents and all benchmark examples is not tractable. Instead, we predict scores for the entire corpus by training an element-wise model (classifier or regressor). We consider different models, such as simple FastText classifiers~\citep{joulin2016fasttext} or fine-tuned language models, which offer tradeoffs between inference speed and prediction accuracy. After scoring, we filter documents by retaining a fixed percentage of tokens (rather than documents) to ensure fair comparisons across methods.

\bydefault{We train a FastText classifier to predict whether documents are in the top 10\% or bottom 90\% by rank, downsampling the majority class to balance the training data (see Section~\ref{sec:main_ablations}). By default, we retain the top 10\% of tokens after scoring and compare different filtering rates in Section~\ref{sec:optimal_filtering}.}

\subsection{Targeting strategies}
\label{sec:targeting_strategies}

BETR can be applied in two ways, depending on whether the goal is optimizing for specific tasks or general capabilities:

\textbf{Evaluation-aware (EA).} When we know what capabilities we want to optimize for, we target their corresponding benchmarks directly. We use equal numbers of examples from each benchmark to prevent larger benchmarks from dominating the selection and target only benchmark training sets to maintain evaluation integrity.

\textbf{Evaluation-blind (EB).} When we want generally capable models, we target many diverse benchmarks while holding out our evaluation suite. We motivate this approach by observing that benchmarks capture human judgments about useful tasks, so targeting diverse benchmarks should select for broadly valuable text. We ensure no overlap between targeting and evaluation benchmarks and do not iterate based on evaluation results (as this would become evaluation-aware).

%% file: figures/main/method_fig.tex
\begin{figure}[t]
  \centering
  \includegraphics[width=\columnwidth]{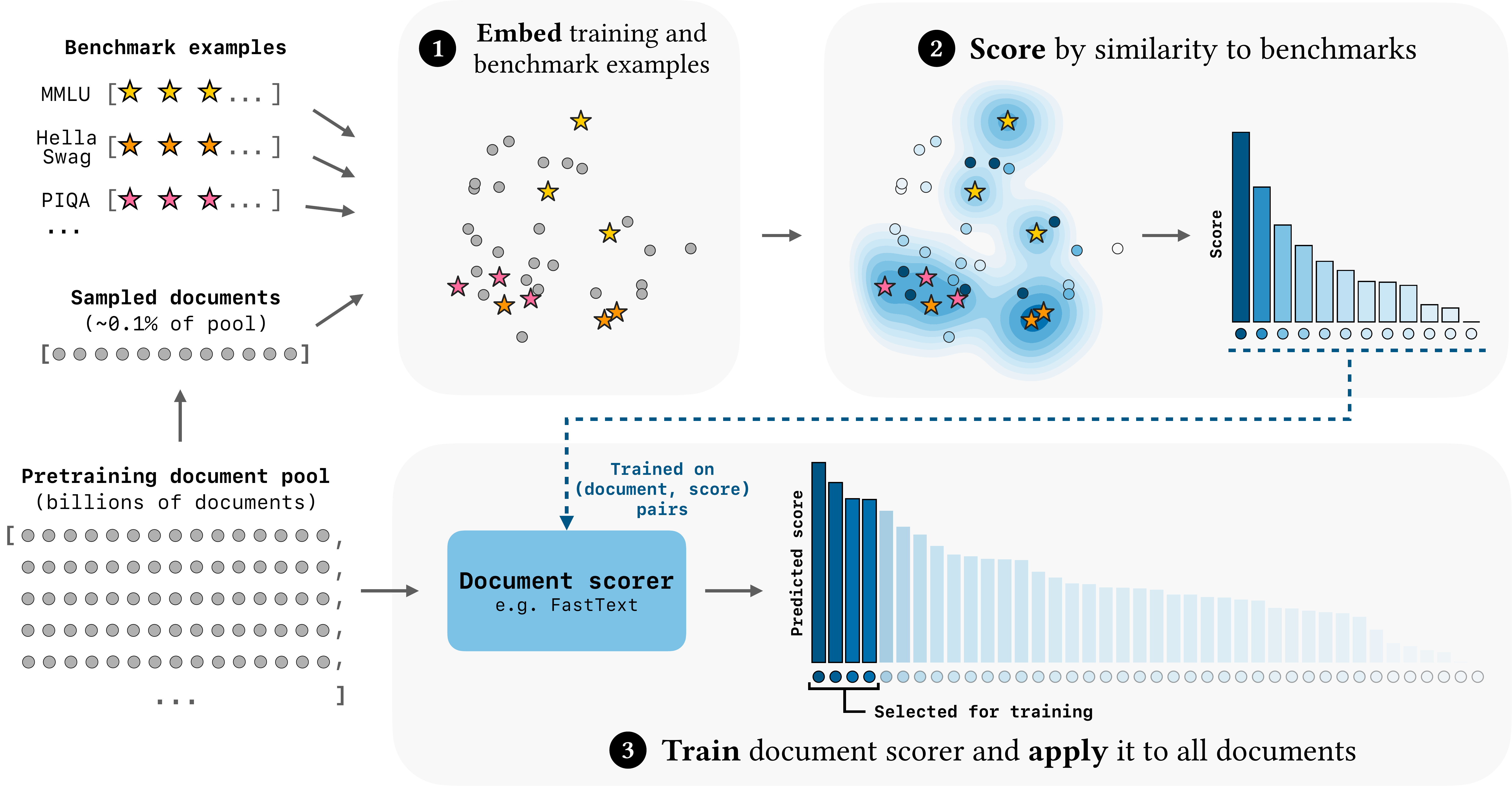}
  \caption{\textbf{BETR method overview.} We embed benchmark examples and a small sample of pretraining documents (\(\sim\)0.1\% of pool) in a shared space, score the sampled documents by their similarity to benchmarks, then train a classifier on these scores to efficiently rank and filter the entire document pool.}
  \label{fig:method}
\end{figure}

%% file: figures/main/rank_sim_distribution.tex
\begin{figure}[t]
    \centering
    \includegraphics[width=\textwidth]{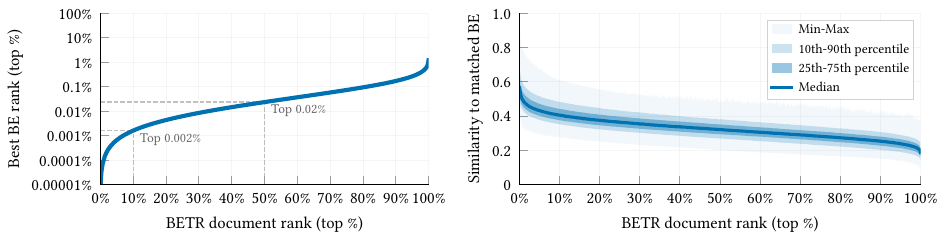}
    \vspace{-1.5em}
    \caption{\textbf{BETR scoring distributions in practice.} 
    \textbf{Left:} Best rank assigned by any benchmark example (\textbf{BE}). With >14,000 benchmark examples competing to score documents, only those ranked in the top 0.002\% by some benchmark example reach the top 10\% of BETR scores. 
    \textbf{Right:} Cosine similarity to the benchmark example that assigned the best rank. Even top-ranked documents show only moderate similarities ($\sim$0.5) to benchmark examples, highlighting the limited overlap between benchmark examples and web text. Shown for our default (i.e. \textcolor{defaultcolor}{``in practice''}) Target-\coremetric settings on DCLM-RefinedWeb.}
    \label{fig:scoring_distributions}
\end{figure}

%% file: sec/4_experimental_setup.tex
\subsection{Benchmarks}

We define two distinct sets of benchmarks, used for both targeting and evaluation.

\textbf{\coremetric benchmarks}. This set consists of MMLU~\citep{hendrycks2020mmlu} and 9 benchmarks from the CoreEN evaluation suite~\citep{gunter2024apple, busbridge2025distillation}: ARC-Easy and ARC-Challenge~\citep{clark2018arc}, HellaSwag~\citep{zellers2019hellaswag}, Lambada OpenAI~\citep{paperno2016lambada}\footnote{Lambada OpenAI only provides a test set. We reserve 1/3 for targeting and evaluate on the remaining 2/3.}, PIQA~\citep{bisk2020piqa}, SciQ~\citep{welbl2017sciq}, TriviaQA~\citep{joshi2017triviaqa}, WebQuestions~\citep{berant2013webqs}, and WinoGrande~\citep{sakaguchi2021winogrande}. These are standard benchmarks that largely overlap with those commonly used in recent data selection work~\citep{li2024dclm,penedo2024fineweb,su2024nemotron,thrush2024improving,liu2024regmix,kong2024large,shum2025predictive}. Targeting \coremetric train sets and evaluating on \coremetric test sets corresponds to our evaluation-aware (EA) setting.

\textbf{\extendedmetric benchmarks}. To test whether benchmark-aligned data generalizes beyond targeted tasks, we construct a diverse benchmark set disjoint from \coremetric. We take DCLM's Core and Extended evaluation suite~\citep{li2024dclm} and remove any benchmarks that overlap with \coremetric (either fully or partially). This gives us 39 benchmarks (detailed in Appendix~\ref{app:benchmark_details}). Targeting \extendedmetric benchmarks while evaluating on \coremetric corresponds to our evaluation-blind (EB) setting.

\subsection{Data pools and baselines}

We run our experiments on two data pools. These are large web-crawl corpora with light filtering that serve as the source for our filtering approaches. For each pool, we compare against its best available filtering method.

\textbf{DCLM-RefinedWeb~\citep{li2024dclm}.}  This data pool processes CommonCrawl through the RefinedWeb pipeline~\citep{penedo2023refinedweb}. The corpus contains 24T tokens, though many near-duplicates remain due to the lack of global deduplication~\citep{tokpanov2024zyda,fang2023data}. Importantly, to ensure evaluation integrity, we apply n-gram decontamination against \coremetric test sets following GPT-3's approach~\citep{brown2020gpt3}. We ablate its impact in Section~\ref{sec:main_ablations} and detail the procedure in Appendix~\ref{app:decontamination}.

\textbf{Nemotron-CC~\citep{su2024nemotron}.} This data pool also processes CommonCrawl but with slightly different preprocessing choices and global fuzzy deduplication. Perhaps most importantly, Nemotron-CC includes 1.9T synthetic tokens created through model-based rephrasing\footnote{Nemotron-CC rephrases documents selectively: after splitting data into quality tiers, the lowest tier is rephrased once while the highest tier is rephrased in 5 different styles. This effectively increases the proportion of ``high-quality'' documents in the dataset.}~\citep{maini2024rephrasing}, resulting in 6.3T total tokens. Since n-gram matching cannot detect rephrased text, we do not decontaminate Nemotron-CC.

\textbf{Baselines.} We compare against two baseline filtering methods. \textbf{DCLM-Baseline}~\citep{li2024dclm} is a FastText classifier trained on Reddit ELI5 + OpenHermes-2.5 as positive examples and RefinedWeb as negatives. When applying it to Nemotron-CC, we retrain the classifier using Nemotron data as negatives instead. For Nemotron-CC, we also compare against \textbf{Nemotron-CC HQ}~\citep{su2024nemotron}, which consists of documents classified as high-quality by an ensemble of classifiers\footnote{We use all documents labeled as high-quality for Nemotron-CC HQ, while \citet{su2024nemotron} only use a subset (actual documents and synthetic QA). Appendix~\ref{app:nemotron_cc_hq_diff} shows their subset performs slightly better, but not enough to change method rankings.}. To ensure fair comparison across methods, we filter all datasets to retain the top 10\% of tokens, except for Nemotron-CC HQ where we use all high-quality documents (as the quality tiers are discrete).

\input{figures/main/scaling_fig}

\subsection{Fixed-scale evaluation}

We first compare datasets using single-scale experiments~\citep{magnusson2025datadecide, li2024dclm}, specifically training 7B parameter models for 140B tokens (the ``7B-1x'' setting in DCLM, roughly compute-optimal per~\citet{hoffmann2022training}). This setting represents the smallest scale at which most datasets surpass random performance on MMLU. After validating our approach through ablations at this scale and confirming consistent improvements via scaling laws, we additionally evaluate our main results at the ``7B-10x'' scale (7B model trained for 1.4T tokens) to test performance in the overtrained regime typical of production models. Further details about our training configuration are provided in Appendix~\ref{app:training}. Since benchmark performance naturally varies, we estimate per-benchmark standard deviation by training 10 models (with different initialization and data shuffling) on a single representative dataset, following T5~\citep{raffel2020exploring}.\footnote{We use variance estimates from the 7B-1x scale for both scales, as estimating them directly at 7B-10x is computationally prohibitive. Actual variance at 7B-10x may differ.}

\subsection{Scaling laws evaluation}
\label{sec:scaling_laws_evaluation}

Single-scale experiments alone cannot determine whether the effects of data selection persist, amplify, or reverse with scale. To address this, we fit scaling laws~\citep{kaplan2020scaling,hoffmann2022training} to compare the performance of different data selection methods across varying compute budgets. We train 53 models per dataset, spanning $6\times10^{18}$ to $6\times10^{21}$ FLOPs (>500 models total). 

Since training loss cannot be used to meaningfully compare different datasets (easier text yields lower loss regardless of quality), we instead predict benchmark performance~\citep{meta2024llama3,gadre2024language,bhagia2024establishing}. This is a two-step process: we first model bits-per-byte for each benchmark as a function of model size and training tokens, then map these to per-benchmark accuracy predictions with uncertainty quantification via bootstrapping (Figure~\ref{fig:scaling_fig}). This enables us to determine which data selection method achieves higher performance at any compute budget and quantify our confidence in these comparisons.

To summarize the relative efficiency of different data selection methods, we report compute multipliers (CM)~\citep{betker_compute_2023,amodei2025deepseek}. A compute multiplier of $X$ between methods A and B means that method A requires only $1/X$ of the compute to achieve the same performance as method B (at compute-optimality). For example, a 2x compute multiplier indicates that one dataset achieves equivalent performance using half the training compute. 

We provide a detailed overview of our scaling law methodology in Appendix~\ref{app:scaling_law_details}.

%% file: figures/main/scaling_fig.tex
\begin{figure}[t]
  \centering
  \includegraphics[width=\columnwidth]{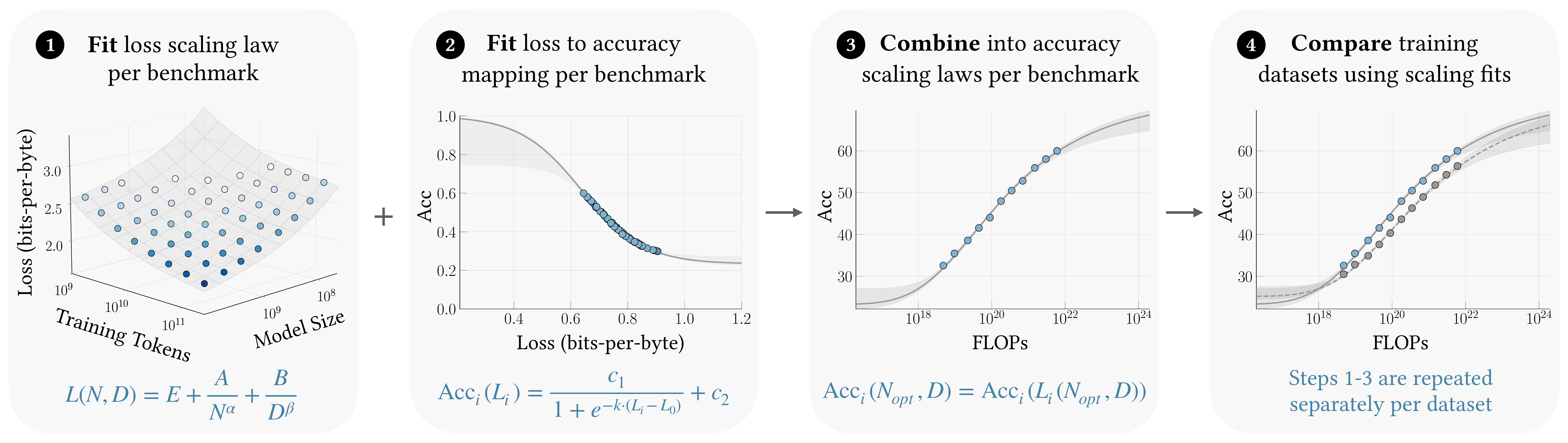}
  \caption{\textbf{Comparing datasets using scaling laws.} For each dataset, we use a two-stage scaling law approach to predict per-benchmark accuracy~\citep{gadre2024language,meta2024llama3,bhagia2024establishing}. 1) Fit a loss scaling law as a function of model size and training tokens (color coding indicates training FLOPs, darker blues for more compute used). 2) Map task loss to accuracy for each benchmark. 3) Combine the two to predict accuracy at any configuration, including compute-optimal settings. 4) Repeat for each dataset to compare performance across compute budgets.}
  \label{fig:scaling_fig}
\end{figure}

%% file: sec/5_results.tex
\subsection{Does directly targeting benchmarks improve performance?}

\input{tables/main/main_results_7b_10x}

We first test whether explicitly aligning pretraining data with target benchmarks improves performance on those same benchmarks. Figure~\ref{fig:pull_figure} shows that BETR Target-\coremetric consistently outperforms all baselines on \coremetric tasks across compute scales from $10^{19}$ to to $10^{22}$ FLOPs (at compute-optimality). On DCLM-RefinedWeb, BETR achieves a 1.8x compute multiplier over DCLM-Baseline (i.e., requires only 55\% of the compute to match DCLM-Baseline performance) and 4.7x over unfiltered data (21\% of compute). On Nemotron-CC, the improvements are even larger: 2.5x over DCLM-Baseline, 2.8x over Nemotron-CC HQ, and 4.7x over no filtering.

These improvements hold across individual benchmarks. BETR Target-\coremetric outperforms all baselines on 9 out of 10 \coremetric tasks on both data pools (all except MMLU on DCLM-RefinedWeb, and all except Winogrande on Nemotron-CC). See Appendix Table \ref{tab:app_efficiency_factors} for per-benchmark compute multipliers.

To verify these improvements persist in the overtrained regime, we evaluate at the 7B-10x scale ($6 \times 10^{22}$ FLOPs). Table~\ref{tab:main_results_10x} shows BETR maintains its advantage on \coremetric: +1.7 points over DCLM-Baseline on DCLM-RefinedWeb, and +1.8 and +1.6 points over Nemotron-CC HQ and DCLM-Baseline respectively on Nemotron-CC.

\keybox{Directly targeting benchmarks achieves a 1.8-2.8x compute multiplier over existing methods (4.7x over no filtering), with consistent improvements across scales and tasks.}

\subsection{Can benchmark targeting precisely control model capabilities?}

\input{tables/main/individual_targeting}

We next test whether BETR enables fine-grained control over model capabilities by targeting each of the 10 \coremetric benchmarks individually, then evaluating on all \coremetric tasks. Table~\ref{tab:individual_targeting} shows clear diagonal dominance: each dataset performs best (or tied best) on its targeted benchmark.

We observe natural clustering within task categories. For example, targeting ARC-Easy improves performance on related knowledge benchmarks like ARC-Challenge, SciQ and MMLU. However, precisely targeting benchmarks in one domain impacts performance in others: e.g., targeting Winogrande improves Lambada (both language understanding benchmarks) but hurts performance on knowledge benchmarks such as MMLU. Targeting all \coremetric benchmarks jointly (bottom row) avoids these trade-offs, ranking 1st to 4th across all tasks.

\keybox{Individual benchmark targeting enables precise control but produces specialized models with narrow capabilities.}

\subsection{Does targeting diverse benchmarks generalize to held-out tasks?}

\input{figures/main/evaluation_blind_joint}

While individual benchmark targeting creates specialized models, we now investigate whether targeting diverse benchmarks can create broadly capable models. We target all 39 \extendedmetric benchmarks (none overlap with \coremetric) and evaluate on \coremetric tasks. This corresponds to our evaluation-blind (EB) setting described in Section~\ref{sec:targeting_strategies}.

We find that BETR Target-\extendedmetric achieves strong performance on held-out \coremetric tasks, matching or exceeding baselines across both scaling laws (Figure~\ref{fig:pull_figure}) and at the 7B-10x scale (Table~\ref{tab:main_results_10x}). As expected, Target-\extendedmetric underperforms Target-\coremetric (which directly targets the evaluation distribution).

To understand why \extendedmetric-targeting works, we use BETR to target ``high-quality'' domains instead (Table~\ref{tab:evaluation_blind_vs_domains}). We test domains from the Pile~\citep{gao2020pile} and the OH-2.5 + ELI5 dataset used as positive examples for DCLM-Baseline's classifier~\citep{li2024dclm}. Performance varies widely depending on the domain: GitHub (52.3) performs worse than no filtering, while DM Mathematics~\citep{saxton2019analysing} (60.4) and OH-2.5 + ELI5 (60.2) approach but never reach Target-\extendedmetric performance (60.9).

We then investigate how performance scales with benchmark diversity (Figure~\ref{fig:evaluation_blind_progression}). Using two random orderings of the 39 \extendedmetric benchmarks\footnote{See Appendix Table~\ref{tab:app_noncore_benchmark_list} for the specific random orderings.}, we measure performance when targeting an increasing number of benchmarks. Performance stagnates from 1 to 5 benchmarks, then reliably improves as more benchmarks are added. This pattern holds across both orderings and data pools.

\keybox{Targeting diverse benchmarks yields generally capable models that perform well even on held-out tasks, with performance improving as more benchmarks are targeted.}

\vspace{-0.5em}

\subsection{What are the consequences of optimizing for a limited set of benchmarks?}

\vspace{-0.5em}

\input{figures/main/core_vs_noncore_fig}

Having shown that diverse benchmark targeting creates broadly capable models, we now test whether optimizing for common evaluation benchmarks maintains general capabilities. We evaluate Target-\coremetric models on held-out \extendedmetric tasks.

Figure~\ref{fig:core_vs_noncore_performance} reveals a clear rank reversal. While Target-\coremetric achieves the best \coremetric performance (66.5 on DCLM-RefinedWeb), it ranks third on \extendedmetric tasks (44.2), trailing both Target-\extendedmetric\footnote{Target-\extendedmetric uses all available benchmark examples without train/test splits or per-benchmark balancing, unlike Target-\coremetric. Our ablations show that train vs test targeting splits has minimal impact (Table~\ref{tab:additional_ablations}).} (46.4) and DCLM-Baseline (46.3). This pattern holds across both data pools.

We analyze document overlap to understand these performance differences (Figure~\ref{fig:document_overlap}).  While methods show moderate agreement (the three methods share 19\% of their top 10\% documents, with pairwise Spearman correlations ranging from 0.44 to 0.71), there is no convergence toward universal ``high-quality'' data. Furthermore, this agreement poorly predicts capability profiles: Target-\extendedmetric and DCLM-Baseline achieve nearly identical performance on both \coremetric and \extendedmetric tasks despite only 32\% agreement, whereas the two BETR variants have 47\% agreement yet produce models with different capabilities.

These findings exemplify Goodhart's Law: despite \coremetric spanning diverse tasks including world knowledge, commonsense reasoning, and language understanding, these benchmarks cease to be good proxies for general capability when directly optimized. The results show that models improve on exactly what their training data targets, with clear trade-offs on other capabilities. This concern extends beyond our work, as many data selection methods evaluate on fewer benchmarks, with some optimizing for as few as 1--2  evaluation tasks~\citep{thrush2024improving, wettig2025organize}.

\keybox{Standard pretraining benchmarks (\coremetric) represent a narrow slice of capabilities: models optimized exclusively for them underperform on held-out tasks.}

\vspace{-0.3em}
\subsection{What factors drive performance?}
\label{sec:main_ablations}

\input{figures/main/ablations}

In Figure~\ref{fig:ablations}, we systematically ablate each BETR component to understand what drives its effectiveness.

\textbf{Target granularity} strongly affects performance (Figure~\ref{fig:ablations}a). Per-example targeting achieves 62.2 \coremetric accuracy, outperforming k-means clustering (60.3--60.6), per-benchmark averaging (59.5), and global averaging (58.3).

\textbf{Embedding choice} has minimal impact (Figure~\ref{fig:ablations}b). Performance is mostly consistent across models, with Arctic-Embed L v2~\citep{yu2024arctic} performing slightly better (62.5)\footnote{We also tested applying a reranker (BGE-v2-m3~\citep{chen2024bge}) to the top 1000 documents per benchmark, which provided no additional benefit.}.

\textbf{The scoring function} significantly affects results (Figure~\ref{fig:ablations}c). Scoring functions that reward documents highly relevant to any single benchmark example (62.2) outperform functions that reward average relevance across multiple examples (58.2).

\textbf{The choice of classifier} has a counterintuitive effect  (Figure~\ref{fig:ablations}d). A simple FastText classifier matches or exceeds LM-based classifiers and regressors despite lower validation accuracy (75.4\% vs 82.4\% for 302M LM), which suggests that classification accuracy is not the key metric for effective data selection.

\input{tables/main/additional_ablations}

\textbf{To verify BETR's integrity}, we test whether gains come from test set information or contamination (Table~\ref{tab:additional_ablations}). Targeting benchmark train sets performs identically to targeting test sets (62.3 vs 62.4), and decontamination has minimal impact (-0.2), confirming the method works without test set leakage.\footnote{For an illustration of what happens when evaluation integrity is \textit{not} maintained, see \citet{schaeffer2023pretraining}.} Finally, directly classifying benchmark vs web text underperforms BETR (59.7 vs 62.2), validating our two-step approach.\footnote{Distinguishing benchmark examples from web text is trivial (99.5\% validation accuracy), while BETR's task of classifying top 10\% vs bottom 90\% of ranked documents is much harder (75.6\% accuracy). We hypothesize that this harder task leads to stronger features for data selection.}

\keybox{Simple design choices drive BETR's success: fine-grained targets, scoring functions that favor single-best matches, and simple classifiers outperform more complex alternatives.}

%% file: tables/main/main_results_7b_10x.tex
\begin{table}[t]
\centering
\begin{adjustbox}{width=\textwidth}
{%
  \setlength{\tabcolsep}{0.5pt}      %
\renewcommand{\arraystretch}{1.0}  %
  \begin{tabular}{
    l@{\hspace{8pt}}    %
    *{10}{G}            %
    @{\hspace{5pt}}c   %
  }
  \toprule
  Method & ARC-E & ARC-C & HellaS & Lamb. & MMLU & PIQA & SciQ & TrQA & WebQs & Wino. & \coremetricavg \\
  \midrule
  \multicolumn{12}{l}{\textbf{Data pool: DCLM-RefinedWeb}} \\ \midrule
  No Filter                          & 77.8 & 47.7 & 60.5 & 74.3 & 54.6 & 80.7 & 94.5 & 43.8 & \textbf{25.6} &\textbf{ 73.2} & 63.3 \\
  DCLM-Baseline   & 81.4 & 50.7 & 60.4 & 75.0 & \textbf{63.1} & 80.3 & 95.9 & 43.4 & 24.4 &\textbf{ 73.4} & 64.8 \\
  BETR Target-\extendedmetric & 81.9 & 51.3 & 58.3 & \textbf{75.5} & \textbf{63.1} & 79.4 & 95.4 & 46.1 & \textbf{25.0} & \textbf{73.8} & 65.0 \\
  \highlight[ontarget]{BETR Target-\coremetric}                    & \textbf{82.9} & \textbf{53.3} & \textbf{63.1} & \textbf{75.8} & \textbf{62.6}& \textbf{83.2} & \textbf{96.7} & \textbf{47.9} & \textbf{26.0} & \textbf{74.0} & \textbf{66.5} \\
  \midrule
  \multicolumn{12}{l}{\textbf{Data pool: Nemotron-CC}} \\ \midrule
  No Filter                          & 82.2 & 51.5 & 60.9 & 71.5 & 60.8 & 80.9 & 95.8 & 45.3 & \textbf{27.6} & 72.1 & 64.9 \\ 
  Nemotron-CC HQ                     & 83.5 & 54.3 & 61.1 & 69.7 & 67.2 & 80.8 & 96.1 & 45.8 & 24.7 & 73.6 & 65.7 \\
  DCLM-Baseline   & 82.4 & 53.5 & 60.8 & 72.3 & 67.3 & 80.4 & \textbf{97.0} & 44.2 & \textbf{27.5} & \textbf{74.7} & 65.9 \\
  BETR Target-\extendedmetric & 83.4 & \textbf{55.1} & 58.7 & 74.0 & \textbf{68.7} & 79.1 & 96.1 & 49.2 & \textbf{26.9} & \textbf{75.2} & 66.6 \\
  \highlight[ontarget]{BETR Target-\coremetric}                 & \textbf{84.0} & \textbf{54.7} & \textbf{64.1} & \textbf{74.9} & \textbf{68.3} & \textbf{83.0} & 96.4 & \textbf{50.2} & 25.5 & \textbf{74.0} & \textbf{67.5} \\
  \bottomrule
  \end{tabular}
}%
\end{adjustbox}
\caption{\textbf{\coremetric results at 7B-10x scale.} We compare BETR variants and baselines at 7B-10x scale (7B parameters, 1.4T tokens). BETR Target-\coremetric \highlight[ontarget]{directly optimizes} for these benchmarks and achieves the highest average performance. BETR Target-\extendedmetric targets benchmarks distinct from \coremetric yet still outperforms baselines, showing generalization beyond targeted tasks. \textcolor{tabgray}{\textbf{Bold}: best result $\pm$ 1 std.}}
\label{tab:main_results_10x}
\begin{flushleft}
\end{flushleft}
\end{table}

%% file: tables/main/individual_targeting.tex
\begin{table}[t]
\centering
\begin{adjustbox}{width=\textwidth}
{%
  \setlength{\tabcolsep}{0.5pt}
  \renewcommand{\arraystretch}{1.0}
  \begin{tabular}{
    l@{\hspace{8pt}}
    *{10}{N}
    @{\hspace{5pt}}c
  }
\toprule
Target & {ARC-E} & {ARC-C} & {HellaS} & {Lamb.} & {MMLU} & {PIQA} & {SciQ} & {TrQA} & {WebQs} & {Wino.} & {\coremetricavg}  \\
\midrule
ARC-Easy      & \cellcolor{ontarget}\textbf{81.1}\rank{1} & 49.8\rank{3} & 58.4\rank{5} & 63.2\rank{8} & 48.1\rank{4} & 81.0\rank{5} & 95.5\rank{2} & 25.7\rank{4} & 11.8\rank{6} & 66.4\rank{8} & 58.1\rank{3} \\
ARC-Challenge & 80.3\rank{3} & \cellcolor{ontarget}\textbf{51.4}\rank{1} & 58.1\rank{6} & 63.3\rank{7} & 48.4\rank{3} & 80.4\rank{6} & 95.3\rank{4} & 25.3\rank{6} &  9.6\rank{9} & 68.7\rank{6} & 58.1\rank{4} \\
HellaSwag     & 74.1\rank{8} & 40.3\rank{8} & \cellcolor{ontarget}\textbf{61.5}\rank{1} & 68.2\rank{4} & 28.7\rank{7} & 82.0\rank{2} & 93.5\rank{7} & 15.6\rank{11} &  7.8\rank{10} & 69.0\rank{5} & 54.1\rank{8} \\
Lambada       & 63.7\rank{11} & 31.1\rank{11} & 53.1\rank{8} & \cellcolor{ontarget}\textbf{75.5}\rank{2} & 26.2\rank{11} & 74.4\rank{9} & 89.6\rank{11} & 25.5\rank{5} & 11.7\rank{7} & 70.6\rank{3} & 52.1\rank{11} \\
MMLU          & 78.6\rank{5} & 47.2\rank{5} & 50.8\rank{9} & 59.2\rank{10} & \cellcolor{ontarget}\textbf{53.0}\rank{1} & 74.1\rank{10} & 95.2\rank{5} & 25.3\rank{7} & 15.6\rank{4} & 69.7\rank{4} & 56.9\rank{6} \\
PIQA          & 75.3\rank{7} & 41.8\rank{7} & 60.5\rank{2} & 65.0\rank{6} & 27.0\rank{10} & \cellcolor{ontarget}\textbf{82.6}\rank{1} & 91.8\rank{8} & 17.3\rank{10} &  6.9\rank{11} & 66.8\rank{7} & 53.5\rank{10} \\
SciQ          & 80.6\rank{2} & \textbf{51.2}\rank{2} & 57.6\rank{7} & 57.9\rank{11} & 47.6\rank{5} & 80.1\rank{7} & \cellcolor{ontarget}\textbf{95.9}\rank{1} & 24.8\rank{8} & 13.3\rank{5} & 66.2\rank{9} & 57.5\rank{5} \\
TriviaQA      & 76.6\rank{6} & 43.7\rank{6} & 50.4\rank{10} & 67.0\rank{5} & 42.2\rank{6} & 75.1\rank{8} & 94.2\rank{6} & \cellcolor{ontarget}\textbf{48.8}\rank{1} & 23.1\rank{2} & 65.7\rank{10} & 58.7\rank{2} \\
WebQs         & 68.8\rank{10} & 34.5\rank{10} & 46.3\rank{11} & 63.1\rank{9} & 28.1\rank{8} & 71.4\rank{11} & 91.7\rank{9} & 41.8\rank{2} & \cellcolor{ontarget}\textbf{27.6}\rank{1} & 64.7\rank{11} & 53.8\rank{9} \\
WinoGrande    & 69.2\rank{9} & 37.0\rank{9} & 59.8\rank{4} & \textbf{75.7}\rank{1} & 27.7\rank{9} & 81.1\rank{4} & 91.4\rank{10} & 21.7\rank{9} & 10.2\rank{8} & \cellcolor{ontarget}\textbf{72.2}\rank{1} & 54.6\rank{7} \\
\midrule
All (i.e. \coremetric) & 79.7\rank{4} & 49.7\rank{4} & 59.9\rank{3} & 74.0\rank{3} & 48.9\rank{2} & 81.3\rank{3} & 95.4\rank{3} & 40.4\rank{3} & 22.5\rank{3} & 71.0\rank{2} & \cellcolor{ontarget}\textbf{62.3}\rank{1} \\
\bottomrule
\end{tabular}
}%
\end{adjustbox}
\caption{\textbf{Individual task targeting.}  Each row shows performance at 7B-1x scale when targeting a single benchmark (or all jointly). Diagonal cells show \highlight[ontarget]{performance on the targeted benchmark}. Individual targeting excels on the targeted task but often reduces performance on others, while joint targeting balances performance across all tasks. \textcolor{tabgray}{\textbf{Bold}: best result $\pm$ 1 std. Superscripts indicate rank within column (1 = best).}}
\label{tab:individual_targeting}
\end{table}

%% file: figures/main/evaluation_blind_joint.tex
\begin{figure}[t]
\begin{minipage}[b]{0.43\textwidth}
\centering
{%
\setlength{\tabcolsep}{0.5em}
\setlength{\abovetopsep}{0pt}
\renewcommand{\arraystretch}{1.0}
\begin{tabular}[t]{lc}
\toprule
Target domain & \coremetricavg \\
\midrule
\textcolor{tabgray}{None} & \textcolor{tabgray}{56.0} \\
GitHub & 52.3 \\
arXiv & 52.6 \\
Wikipedia (en) & 55.9 \\
StackExchange & 56.3 \\
OH-2.5 + ELI5 & 60.2 \\
DM Mathematics & 60.4 \\
All above domains & 59.6 \\
\midrule
\extendedmetric benchmarks & \textbf{60.9} \\
\bottomrule
\end{tabular}
}
\vspace{0.2em}
\captionof{table}{\textbf{Benchmark vs. domain targeting.} Targeting diverse benchmarks (\extendedmetric) outperforms targeting specific domains on held-out tasks (\coremetric).}
\label{tab:evaluation_blind_vs_domains}
\end{minipage}
\hfill
\begin{minipage}[b]{0.54\textwidth}
\centering
\raisebox{-\height}{%
    \includegraphics[width=\textwidth]{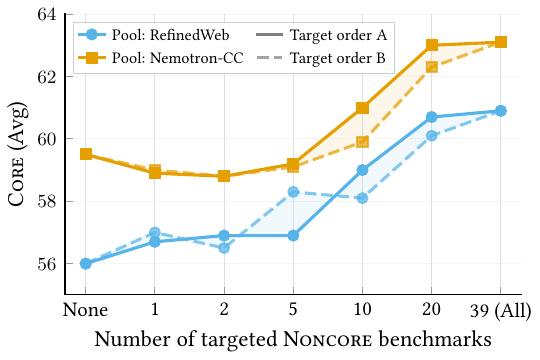}
}
\captionof{figure}{\textbf{Performance scales with benchmark diversity.} After an initial plateau, targeting additional diverse benchmarks (\extendedmetric) consistently improves performance on held-out tasks (\coremetric).}
\label{fig:evaluation_blind_progression}
\end{minipage}
\end{figure}

%% file: figures/main/core_vs_noncore_fig.tex
\begin{figure}[t]
\begin{minipage}[t]{0.54\textwidth}
    \centering
    \includegraphics[width=\textwidth]{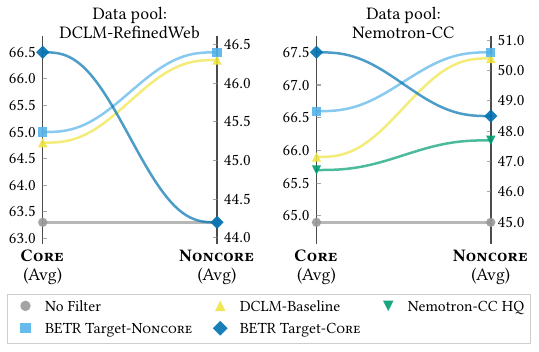}
    \vspace{-0.3em}
    \caption{\textbf{\coremetric vs \extendedmetric performance.}
    At 7B-10x scale, BETR Target-\coremetric achieves the highest \coremetric performance but falls to third place on held-out \extendedmetric tasks, showing the trade-off of targeted optimization.}
    \label{fig:core_vs_noncore_performance}
\end{minipage}
\hfill
\begin{minipage}[t]{0.44\textwidth}
    \centering
    \includegraphics[width=\textwidth]{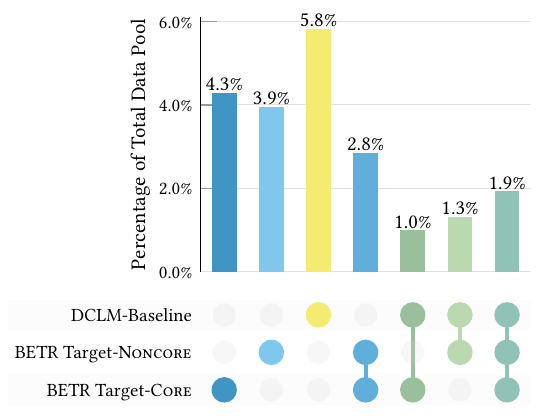}
    \vspace{-0.3em}
    \caption{\textbf{Document overlap analysis.}  Top 10\% document agreement on DCLM-RefinedWeb poorly correlates with model capabilities: similar-performing methods show different selection patterns.}
    \label{fig:document_overlap}
\end{minipage}
\end{figure}

%% file: figures/main/ablations.tex
\begin{figure*}[t]
\centering
\begin{minipage}[t]{0.54\textwidth}
    \centering
    (a) Target granularity\\[0.0em]
    \includegraphics[width=\textwidth]{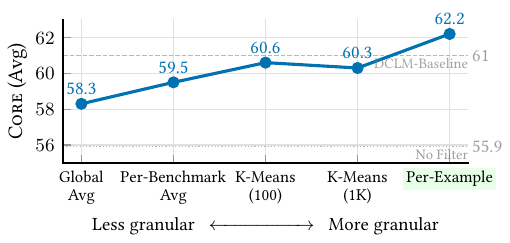}
\end{minipage}
\hfill
\begin{minipage}[t]{0.43\textwidth}
    \centering
    (b) Embedding model\\[0.8em]
    \small
    {%
    \setlength{\tabcolsep}{0.6em}        %
    \renewcommand{\arraystretch}{1.1}  %
    \begin{tabular}{lc}
    \toprule
    Embedding model & \coremetricavg \\
    \midrule
    GTE-Base v1.5 & 61.9 \\
    GTE-Large v1.5 & 62.2 \\
    GTE-Large v1.5 + Reranker & 62.2 \\
    GTE-Qwen2 1.5B & 60.1 \\
    \highlight[selected]{Arctic-Embed L v2} & 62.5 \\
    \bottomrule
    \end{tabular}
    }
\end{minipage}

\vspace{0.8em} %

\begin{minipage}[t]{0.54\textwidth}
    \centering
    (c) Scoring function\\[0.0em]
    \includegraphics[width=\textwidth]{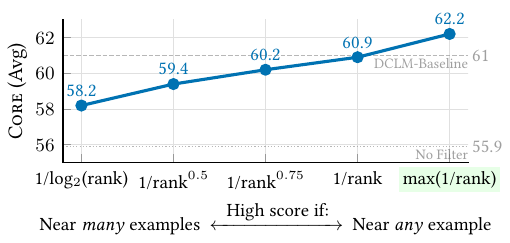}
\end{minipage}
\hfill
\begin{minipage}[t]{0.43\textwidth}
    \centering
    (d) Elementwise scoring model (DFN)\\[0.6em]
    \small
     {%
     \setlength{\tabcolsep}{0.4em}        %
     \renewcommand{\arraystretch}{1.1}  %
    \begin{tabular}{lc>{\color{gray}}c}
    \toprule
    DFN & \coremetricavg & Cls. acc. \\
    \midrule
    \highlight[selected]{FastText} & 62.2 & 75.4 \\
    LM 85M (Class.) & 61.8 & 79.8 \\
    LM 302M (Class.) & 61.6 & 82.4 \\
    \midrule
    LM 85M (Reg.) & 61.5 & --- \\
    LM 302M (Reg.) & 61.9 & --- \\
    \bottomrule
    \end{tabular}
    }
\end{minipage}
\vspace{-0.3em}
\caption{\textbf{BETR ablation studies} showing the impact of (a) target granularity, (b) embedding model choice, (c) scoring function, and (d) elementwise scorer on model performance. All experiments use DCLM-RefinedWeb at 7B-1x scale. \highlight[selected]{Highlighted cells} indicate selected settings for our main BETR models in Figure~\ref{fig:pull_figure} and Table~\ref{tab:main_results_10x}.}
\label{fig:ablations}
\end{figure*}

%% file: tables/main/additional_ablations.tex
\begin{table*}[t]
\centering
\begin{minipage}[t]{0.36\textwidth}
    \centering
    \small
     {%
    \setlength{\tabcolsep}{0.2em}        %
    \begin{tabular}{lc}
    \toprule
    Benchmark target & \coremetricavg \\
    \midrule
    Test set (i.e. eval) & 62.4  \\
    Train set (i.e. non-eval) & 62.3\\
    \bottomrule
    \end{tabular}
    }  
\end{minipage}
\hfill
\begin{minipage}[t]{0.21\textwidth}
    \centering
    \small
    {%
    \setlength{\tabcolsep}{0.4em}        %
    \begin{tabular}{lc}
    \toprule
    Decont. & \coremetricavg  \\
    \midrule
    \ding{55} & 62.5  \\
    \ding{51} & 62.3  \\
    \bottomrule
    \end{tabular}
    }
\end{minipage}
\hfill
\begin{minipage}[t]{0.38\textwidth}
    \centering
    \small
    {%
    \setlength{\tabcolsep}{0.2em}        %
    \begin{tabular}{lc>{\color{gray}}c}
    \toprule
    Method & \coremetricavg & Cls. acc. \\
    \midrule
    \coremetric vs crawl  & 59.7 & 99.5 \\
    BETR Target-\coremetric & 62.3 & 75.6 \\
    \bottomrule
    \end{tabular}
    }
\end{minipage}
\vspace{-0.2em}
\caption{\textbf{Method integrity ablations} showing (left) targeting train vs test splits yields similar results, (middle) decontamination has minimal impact, and (right) BETR significantly outperforms directly training a FastText classifier on benchmarks against web crawl.}
\label{tab:additional_ablations}
\end{table*}

%% file: sec/6_scaling_law_insights.tex
Our scaling analysis reveals additional insights about data selection that only become visible across multiple compute scales. While Section~\ref{sec:results} showed that BETR consistently improves \coremetric performance, examining individual benchmarks and filtering rates reveals important nuances: not all tasks benefit equally from data selection, and optimal filtering strategies must adapt to model scale.

\subsection{Tasks vary in how much they benefit from data selection}

\input{figures/main/task_acc_scaling_comparison}

We test whether all benchmarks benefit equally from data selection. While BETR achieves a 4.7x average compute multiplier over no filtering, examining individual benchmarks reveals whether some tasks are easier to improve through data selection.

Figure~\ref{fig:nemotron_task_acc_scaling_comp} shows per-benchmark FLOPs-to-accuracy curves for BETR Target-\coremetric and no filtering on Nemotron-CC. From these curves, we find substantial variation in the effectiveness of data selection: BETR achieves a 10.4x compute multiplier over no filtering on ARC-Easy and 9.5x on PIQA (i.e., requires 10\% of the FLOPs to match no filtering performance), but only 2.8x on WebQuestions and 1.8x on Winogrande. MMLU falls in the middle at 5.2x. See Appendix Table~\ref{tab:app_efficiency_factors} for complete results.

This 6-fold range shows that data selection benefits tasks unevenly, and suggests matching works best when benchmark-relevant information is concentrated in specific documents (e.g., knowledge-intensive benchmarks) rather than distributed across general text (e.g., language understanding benchmarks). All tasks improve with data selection, but some see order-of-magnitude efficiency gains while others benefit primarily from scale.

\keybox{Data selection benefits tasks unevenly. For BETR Target-\coremetric, knowledge tasks achieve up to 10x compute multipliers (ARC-Easy) while language understanding tasks show more modest 2x multipliers (Winogrande).}

\subsection{Optimal filtering rate depends strongly on scale}
\label{sec:optimal_filtering}
\input{figures/main/optimal_filtering_rate_scaling}

Throughout our experiments, each method retains the top 10\% of tokens to ensure fair comparisons. However, BETR produces continuous scores that can be used to rank and filter documents at any threshold. We therefore test whether 10\% is optimal across scales by training models with BETR Target-\coremetric on Nemotron-CC at four filtering rates: top 3\%, 10\%, 30\%, and 100\% (no filtering).

Figure~\ref{fig:optimal_filtering_scaling_main} reveals a clear trend across scales. The left panel shows \coremetric average accuracy curves for each filtering rate: aggressive top 3\% filtering achieves the highest accuracy at small compute budgets but is overtaken by progressively lighter filtering as compute increases. The middle panel quantifies these transitions using bootstrap distributions from our scaling fits. The probability of each rate being optimal shifts smoothly from 3\% (below $10^{21}$ FLOPs) to 10\% (around $10^{22}$ FLOPs) to 30\% (beyond $10^{23}$ FLOPs). The right panel shows this progression follows a power law: $F_{\text{opt}}(C) = 4 \times 10^{-5} \times C^{0.25}$, where $F_{\text{opt}}$ is the percentage of tokens retained and $C$ is training FLOPs.\footnote{We fit a power law for its simplicity and good fit in our observed range. Other functional forms (e.g., sigmoid) might better capture behavior at extreme scales where filtering approaches 100\%, but we do not reach such scales in our experiments.} These results hold at compute-optimal training. However, when training fixed-size models for increasing token counts (Appendix~\ref{app:subsection_fixed_size}), the shift toward lighter filtering is much more gradual.

While \citet{goyal2024scaling} observed similar scale-dependent filtering for CLIP training, their results emerge in data-constrained settings with repeated epochs. Our findings show this pattern holds even in single-epoch language model training, suggesting smaller models inherently require more selective filtering while larger models can extract signal from noisier but more diverse data. Similar patterns appear on DCLM-RefinedWeb (Appendix~\ref{app:subsection_dclm_opt_filterin}), indicating this is not data pool-specific.

\keybox{Optimal filtering predictably becomes less strict as model scale increases. For BETR Target-\coremetric on Nemotron-CC, the optimal filtering rate (top-$x\%$) follows $F_{\text{opt}} \propto C^{0.25}$, increasing from 3\% at $10^{20}$ FLOPs to 30\% at $10^{23}$ FLOPs.}

%% file: figures/main/task_acc_scaling_comparison.tex
\begin{figure}[t]
\centering
\includegraphics[width=\columnwidth]{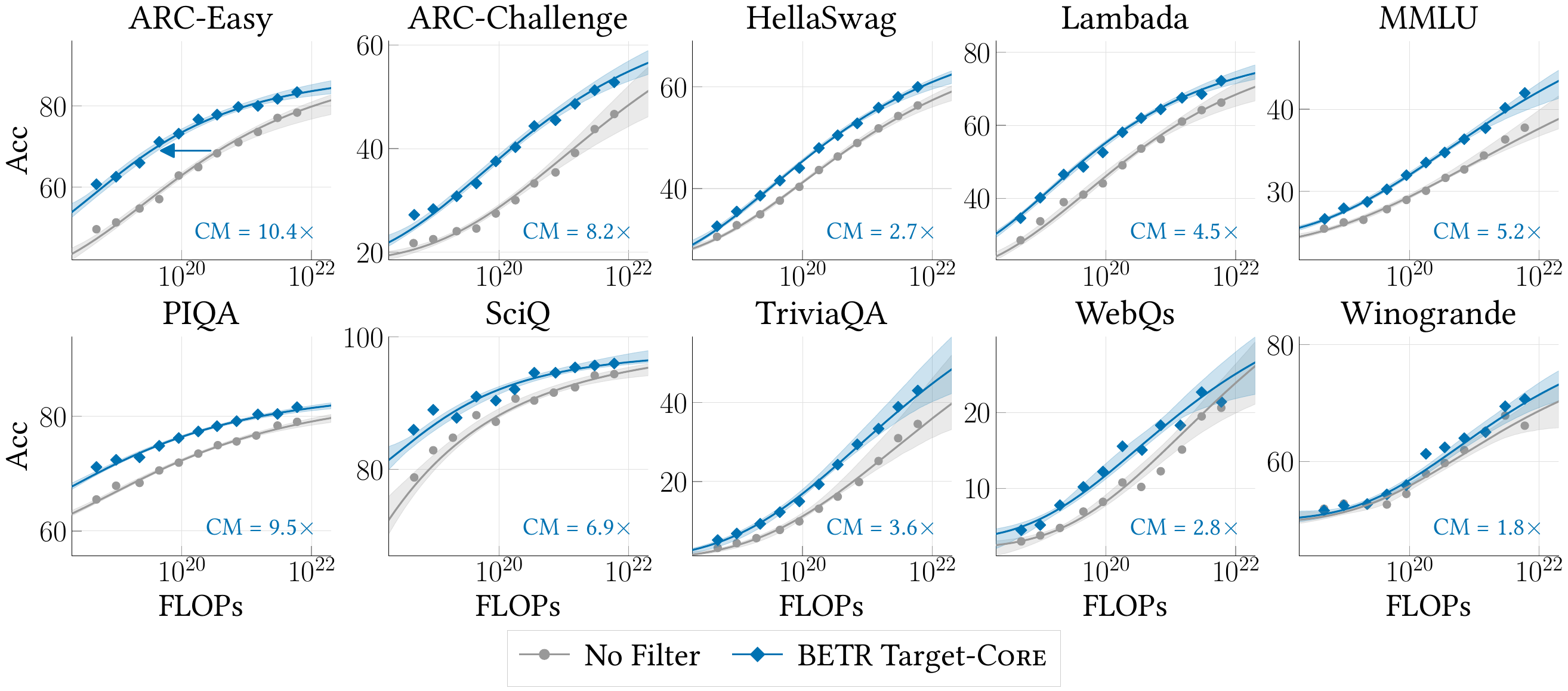}

\caption{\textbf{Per-benchmark scaling laws.} We compare \textcolor{wongBlue}{BETR Target-\coremetric} against \textcolor{wongGrey}{no filtering} across all \coremetric benchmarks on Nemotron-CC, showing compute-optimal performance from $10^{19}$ to $10^{22}$ FLOPs. The scaling laws use two-stage fitting: first from model size and training tokens to per-benchmark bits-per-byte, then from bits-per-byte to accuracy. Compute multipliers (CM) vary by over 5x: from 10.4x for ARC-Easy to 1.8x for Winogrande.}
\label{fig:nemotron_task_acc_scaling_comp}
\end{figure}

%% file: figures/main/optimal_filtering_rate_scaling.tex
\begin{figure}[t]
\centering
\includegraphics[width=\columnwidth]{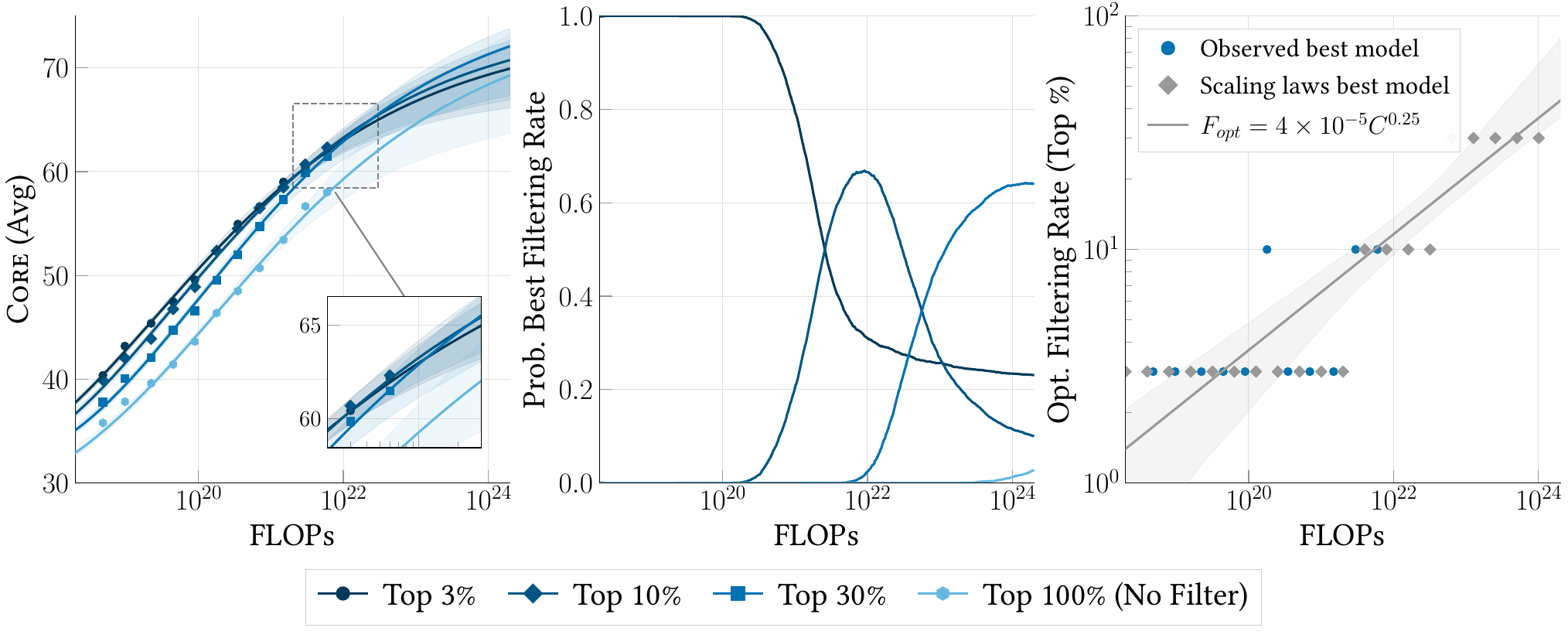}
\caption{\textbf{Optimal data filtering rate scaling.} \textbf{Left:} Compute-optimal \coremetric accuracy for varying BETR Target-\coremetric filtering rates on Nemotron-CC, along with scaling law fits. \textbf{Center:} Probability that each filtering rate is optimal at a given compute scale, estimated from bootstrap distributions of scaling law fits. \textbf{Right:} Optimal filtering rate (top-$x\%$) as a function of compute, shown both for the \textcolor{wongBlue}{best observed model} and the \textcolor{gray}{best model predicted via scaling laws}. The scaling law fit $F_{\mathrm{opt}}(C) = 4 \times 10^{-5} C^{0.25}$ is overlaid for comparison.}
\label{fig:optimal_filtering_scaling_main}
\end{figure}

%% file: sec/7_discussion_conclusion.tex
We tested whether language models improve when pretraining data matches target tasks. This hypothesis seems almost self-evident: training on relevant data should naturally improve relevant capabilities. Yet, quantifying this effect and understanding its implications provides important insights into data selection and evaluation practices. When targeting evaluated benchmarks, BETR achieves 1.8--2.8x compute multipliers over existing methods, with consistent improvements from $10^{19}$ to $10^{23}$ FLOPs. Additionally, the method works equally well for creating specialists (via individual benchmark targeting) or generalists (via diverse benchmark targeting), making it a valuable tool for explicit data selection.

Beyond its practical utility, BETR also sheds light on important properties of benchmark optimization. Target-\coremetric achieves the best \coremetric performance but falls below baselines on held-out \extendedmetric tasks. In contrast, Target-\extendedmetric performs best on \extendedmetric while maintaining competitive \coremetric performance. This asymmetry shows BETR works as intended: each variant excels where targeted. It also highlights that \coremetric benchmarks represent a narrower slice of capabilities than their diversity (world knowledge, language understanding, commonsense reasoning) would suggest. This pattern extends beyond BETR: existing methods implicitly optimize for similar benchmarks through iterative development, with BETR simply accelerating this process to reveal its endpoint.\footnote{This reflects Goodhart's Law: when a measure becomes a target, it ceases to be reliable. \coremetric benchmarks, initially proxies for general language capabilities, lose their indicative power once explicitly optimized.} Achieving broader capabilities thus requires targeting benchmarks beyond those commonly used for evaluation.

Although explicit targeting might seem at odds with pretraining’s traditional emphasis on generality, our scaling analysis offers a reconciling insight: as compute increases, optimal filtering becomes predictably less strict. Smaller models perform best when trained on narrowly filtered datasets, while larger models benefit from more diverse data. At sufficient scales, lighter filtering combined with diverse targeting could resolve the specialization/generality trade-off.

Our findings have several limitations. We experiment only with English text, excluding multilingual and code data, where BETR may behave differently.\footnote{Applying BETR to multilingual or code data would likely require replacing FastText's whitespace tokenization with a BPE tokenizer~\citep{zhu2024deepseekcoderv2}.} We do not verify if pretraining improvements persist after post-training.\footnote{\citet{li2024dclm} suggest such improvements typically persist.} Our scaling analysis assumes unlimited data, while aggressive filtering realistically leads to data constraints~\citep{muennighoff2023scaling}.\footnote{For instance, top 10\% filtering of Nemotron-CC yields only 630B tokens, far below the 15T+ tokens typical in production models~\citep{meta2024llama3,deepseek2024v3,qwen2025v3}.} We also focus exclusively on hard filtering at fixed thresholds, rather than softer weighting or multi-stage approaches~\citep{blakeney2024does}. Finally, FastText's strong performance despite lower classification accuracy than LM alternatives remains unexplained, indicating gaps in our understanding of data selection.

Taken together, our findings demonstrate that explicit data-task matching is both feasible and useful, achieving 2--3x compute multipliers over existing methods while enabling precise control over capabilities. By making data selection targets explicit, BETR reveals that benchmark optimization (whether deliberate or implicit) shapes model capabilities in predictable ways. These findings suggest that progress in language modeling requires not just better data selection methods, but clarity about what capabilities we want and how we measure them.

\textbf{Acknowledgments.} We would like to thank Roman Bachmann, Sam Dodge, Alaaeldin El-Nouby, Doug Kang, Oğuzhan Fatih Kar, Xiang Kong, Justin Lazarow, Brandon McKinzie, Futang Peng, Henry Tran, and Yinfei Yang for helpful feedback and support at various stages of the project.

%% file: sec/8_appendix.tex
\section*{\LARGE Appendix}
\section*{Table of Contents}
\startcontents[appendices]
\printcontents[appendices]{l}{1}{\setcounter{tocdepth}{2}}
\newpage

\section{Method details}

We briefly describe additional implementation details for BETR.

\textbf{Embedding documents.} Pretraining documents are embedded directly without any preprocessing. For benchmark examples, we convert each example into a single document by concatenating the question, answer, and any additional information (e.g., SciQ~\citep{welbl2017sciq} includes a ``support'' field with explanatory text for the correct answer). The maximum context length is set to 8192 tokens for both Arctic-Embed L v2~\citep{yu2024arctic} and GTE models~\citep{li2023gte,zhang2024mgte}.

\textbf{FastText scorer training.} All FastText models~\citep{joulin2016fasttext} use whitespace tokenization and 2-grams with the following hyperparameters: learning rate of 0.03, dimension of 128, window size of 10, minimum occurrence count of 5, and 5 training epochs. We selected these values through 1000 Optuna trials~\citep{akiba2019optuna} on a representative baseline configuration, observing that validation accuracy remained stable across a wide range of hyperparameters near these values.

\textbf{Language model scorer training.} For the LM-based classification and regression scorers~\citep{kong2024large} in Figure~\ref{fig:ablations}, we finetune pretrained language models with either a binary classification or regression head. We select hyperparameters through $\sim$50 runs, searching over training steps, learning rates, schedules, and whether to freeze the base model.

\textbf{Filtering thresholds.} We determine filtering thresholds by applying scorers to a held-out set of 100K documents and selecting thresholds that retain the desired percentage of tokens (typically 10\%). We find that this sample size is sufficient, as thresholds stabilize at 100K documents.

\section{Experimental setup details}

\subsection{Benchmark details}
\label{app:benchmark_details}

\textbf{\coremetric evaluation settings.} Our \coremetric evaluation settings follow from \citet{gunter2024apple}. ARC-Easy, ARC-Challenge~\citep{clark2018arc}, HellaSwag~\citep{zellers2019hellaswag}, Lambada OpenAI~\citep{paperno2016lambada}, PIQA~\citep{bisk2020piqa}, SciQ~\citep{welbl2017sciq}, WinoGrande~\citep{sakaguchi2021winogrande} are 0-shot tasks. TriviaQA~\citep{joshi2017triviaqa} and WebQuestions~\citep{berant2013webqs} are 1-shot tasks. MMLU~\citep{hendrycks2020mmlu} is a 5-shot task.

\textbf{\extendedmetric evaluation settings.} Our \extendedmetric evaluation tasks are sourced from the DCLM-Core and DCLM-Extended evaluation sets~\citep{li2024dclm}, taking all tasks that do not (partially) overlap with \coremetric tasks. The evaluation procedure also follows from \citet{li2024dclm}. The benchmarks are all listed in Table~\ref{tab:app_noncore_benchmark_list}, along with their subcategory from Table~\ref{tab:app_noncore_results_10x} and random order from Figure~\ref{fig:evaluation_blind_progression}.

\subsection{Decontamination}
\label{app:decontamination}

We decontaminate DCLM-RefinedWeb using the procedure from GPT-3~\citep{brown2020gpt3} and MT-NLG~\citep{smith2022using}, as implemented in NeMo-Curator~\citep{jennings2025nemocurator}. This procedure uses n-gram matching (with n ranging from 8 to 13) to identify overlaps between pretraining data and \coremetric benchmarks. When a matching n-gram is found, up to 200 characters on each side are removed and the document is split at that point. Documents that get split more than 10 times are discarded entirely. To preserve common phrases, we skip decontamination for n-grams appearing more than 10,000 times in the pretraining dataset. This provides a large safety margin, as we manually observe that the transition between benchmark content and common phrases is around 300 occurrences.

\subsection{Fixed-scale architecture}
\label{app:training}

The model used for fixed-scale experiments (7B-1x and 7B-10x scale) is a 6.7B Transformer decoder~\citep{vaswani2017attention}. The architecture follows PaLM~\citep{chowdhery2023palm} and Llama3~\citep{meta2024llama3}, i.e., it uses SwiGLU~\citep{shazeer2020swiglu}, GQA~\citep{ainslie2023gqa}, RMSNorm~\citep{zhang2019rms}, and RoPE~\citep{su2024rope}. We also use $\mu$P-simple~\citep{yang2022tensor, wortsman2023small} and QKNorm~\citep{wortsman2023small}. The model has $n_{\text{layers}} = 32$, $d_{\text{model}} = 4096$, $d_{\text{ff}} = 13056$, $n_{\text{query}} = 32$, $n_{\text{kv}} = 8$, and a context length of 4096. We use AXLearn~\citep{lee2025axlearn} as our training framework.

\input{sec/8_app_scaling}

\clearpage

\input{sec/8_app_extra_results}

%% file: sec/8_app_scaling.tex
\section{Scaling law methodology}
\label{app:scaling_law_details}

This section of the appendix provides complete details of our scaling law methodology, which enables us to compare data selection methods across compute scales and derive the compute multipliers reported in Sections \ref{sec:results} and \ref{sec:scaling_insights}.

\subsection{Scaling experiments setup}

\textbf{Overview.} To robustly compare how different data selection methods perform across scales, we trained over 500 models spanning 11 different datasets (i.e. data pool + selection method). This large-scale experiment enables us to fit reliable scaling laws that predict performance across compute budgets from $10^{19}$ to $10^{22}$ FLOPs.

\textbf{Experimental grid.} We systematically vary two key dimensions, using 53 model configurations spanning:
\begin{itemize}
\item \textbf{Model sizes}: 50M, 91M, 175M, 343M, 790M, 1.6B, 3.1B, 6.6B parameters
\item \textbf{Training tokens}: 1.1B, 2.3B, 4.7B, 9.4B, 19B, 38B, 76B, 152B tokens
\end{itemize}

This grid provides comprehensive coverage of the (model size, training tokens) parameter space to allow fitting the Chinchilla-style scaling laws described in Equation \ref{eq:loss_scaling_law}.

\textbf{Training configuration.} All models use an architecture similar to the one described in Section~\ref{app:training}, but without grouped-query attention (GQA) for simplicity. Model width and depth vary to match the desired parameter count, and the learning rate is set to $10^{-2}$ for all models as we use $\mu$P-simple. After training, we measure the pretraining dataset validation loss, benchmark accuracy, and bits-per-byte (BPB) on the golden answers for each benchmark. These measurements form the basis of our two-stage prediction pipeline: first predicting BPB from model size and training tokens, then mapping BPB to task accuracy.

\textbf{MMLU reformulation for stable scaling.} In standard MMLU evaluation~\citep{hendrycks2020mmlu}, models are shown multiple-choice questions and must answer by reference (i.e., select from options labeled A, B, C, or D). This format produces inconsistent and unpredictable performance patterns across different model sizes~\citep{schaeffer2024has}, making it difficult to forecast how larger models will perform. Following \citet{bhagia2024establishing}, we reformulate MMLU by removing the answer-by-reference requirement. Instead, we present only the question and evaluate the model's log-likelihood of each option text directly. We call this variant ``MMLU-Direct''.

\begin{figure}[!h]
\centering
\includegraphics[width=\columnwidth]{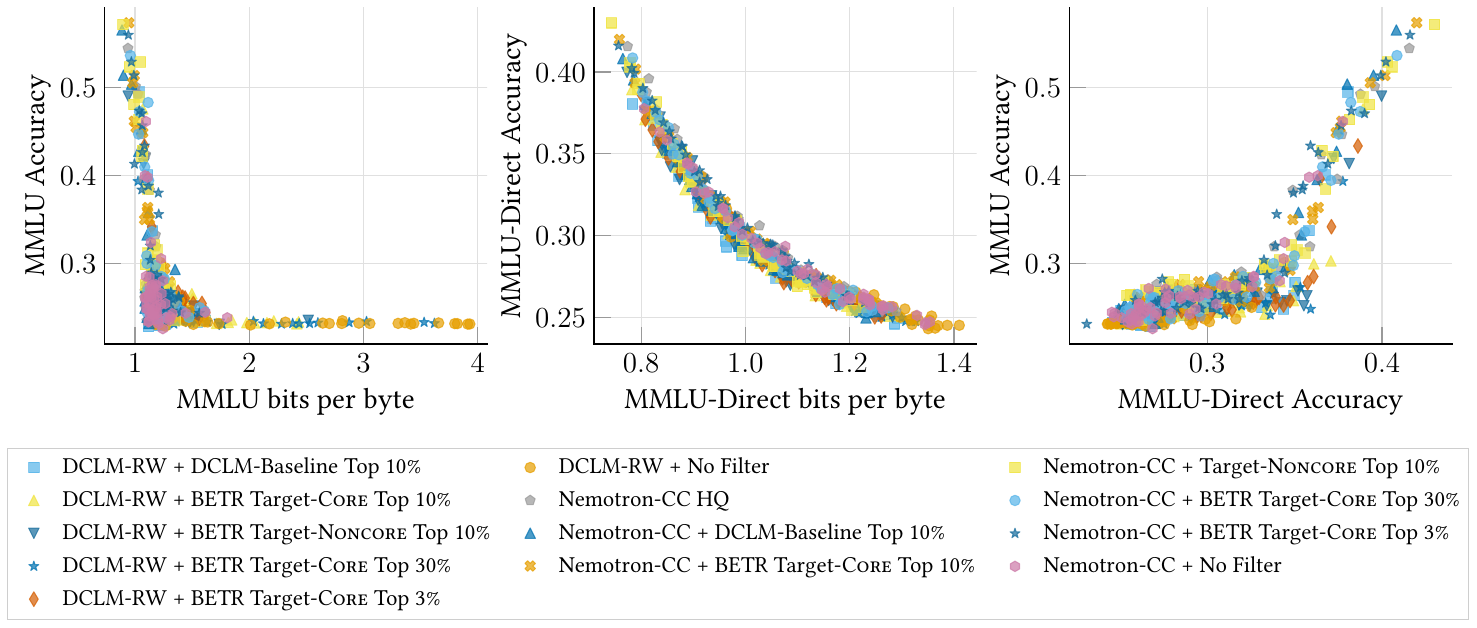}
\caption{\textbf{MMLU task accuracy vs. bits-per-byte scaling.} Comparison between the original MMLU evaluation format (i.e. by reference)and the reformulated MMLU-Direct variant. \textbf{Left:} Original MMLU accuracy as a function of bits-per-byte shows irregular patterns. \textbf{Middle:} MMLU-Direct accuracy as a function of bits-per-byte shows a clean, predictable relationship. \textbf{Right:} Direct comparison of accuracy between the original and reformulated versions, showing a strong monotonic relationship. We use MMLU-Direct in all scaling law analyses for its cleaner scaling properties.}
\label{fig:app_mmlu_vs_mmlu_direct}
\end{figure}

Figure~\ref{fig:app_mmlu_vs_mmlu_direct} shows the impact of this reformulation. The original (i.e. by reference) MMLU format yields irregular accuracy-to-BPB relationships (left), while MMLU-Direct shows smooth, predictable scaling (center). The two formats remain strongly correlated (right), confirming that MMLU-Direct preserves the benchmark's discriminative power while enabling cleaner scaling analysis. However, MMLU-Direct typically produces lower absolute accuracy scores for models above $10^{21}$ FLOPs. All subsequent references to MMLU in our scaling analysis use this reformulated version.

\subsection{Batch size selection}

\begin{figure}[!h]
\centering
\includegraphics[width=0.55\textwidth]{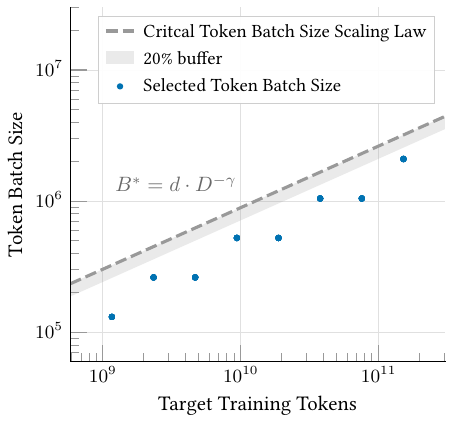}
\caption{\textbf{Batch size selection for scaling experiments.} The dashed gray line shows the critical batch size $B^*$ as a function of training tokens following \citet{zhang2024does}. The light gray region indicates our 20\% safety buffer below $B^*$. Blue circles mark our selected batch sizes, which are rounded down to the nearest power of 2 from the buffer threshold. This ensures all models train below their critical batch size.}
\label{fig:app_batch_size_scaling}
\end{figure}

Large batch sizes hurt model performance beyond a critical threshold \citep{mccandlish2018empirical}. For fair scaling comparisons, we must ensure all models train below this threshold. We follow the critical batch size scaling law from \citet{zhang2024does}, which found that the critical token batch size $B^*$ depends primarily on total training tokens $D$:

\begin{equation}
    B^* = d \cdot D^{-\gamma},
\end{equation}

where $\gamma = -0.47$ and $d = 22.91$ based on their empirical measurements. To remain safely below the critical threshold, we apply a 20\% reduction from $B^*$, then round down to the nearest power of 2 for implementation convenience. Figure~\ref{fig:app_batch_size_scaling} illustrates this selection procedure across our range of training tokens (1B to 152B).

\FloatBarrier

\subsection{Loss scaling with model size and training tokens}

With all models trained and evaluated, we now fit scaling laws to predict loss at arbitrary compute budgets. This forms the first stage of our two-stage prediction pipeline.

We fit scaling laws of the form:
\begin{equation}
L_i(N, D) = E + \frac{A}{N^\alpha} + \frac{B}{D^\beta},
\label{eq:loss_scaling_law}
\end{equation}

where $L_i$ is either validation loss or benchmark-specific BPB, $N$ is model size (total parameters), $D$ is training tokens, and $A$, $B$, $E$, $\alpha$, $\beta$ are fitted parameters. This functional form, introduced by \citet{hoffmann2022training}, captures how loss improves with both model size and training tokens.

Following \citet{besiroglu2024chinchilla}, we fit these parameters by minimizing the Huber loss between predicted and observed values:

\begin{equation}
\min_{a, b, e, \alpha, \beta} \sum_{\text{Run } i} 
\mathrm{Huber}_\delta \left( 
    \mathrm{LSE}\left(
        a - \alpha \log N_i,\, 
        b - \beta \log D_i,\, 
        e
    \right) 
    - \log L_i
\right) ,
\end{equation}

where LSE denotes log-sum-exp and $\delta = 10^{-3}$. To improve numerical stability and ensure positivity of parameters, we reparameterize the optimization in log space as $a = \log A$, $b = \log B$, and $e = \log E$, and directly optimize over ($a$, $b$, $\alpha$, $\beta$, $e$). We use the BFGS optimizer implemented in {\small\texttt{scipy.optimize.minimize}}, beginning with a grid search over initial parameter values to identify good starting points. Finally, we perform 4000 bootstrap resamplings of the training data points (with replacement) to generate a distribution over fitted curves, enabling uncertainty estimation in downstream predictions.

\begin{figure}[!h]
\centering
\includegraphics[width=\columnwidth]{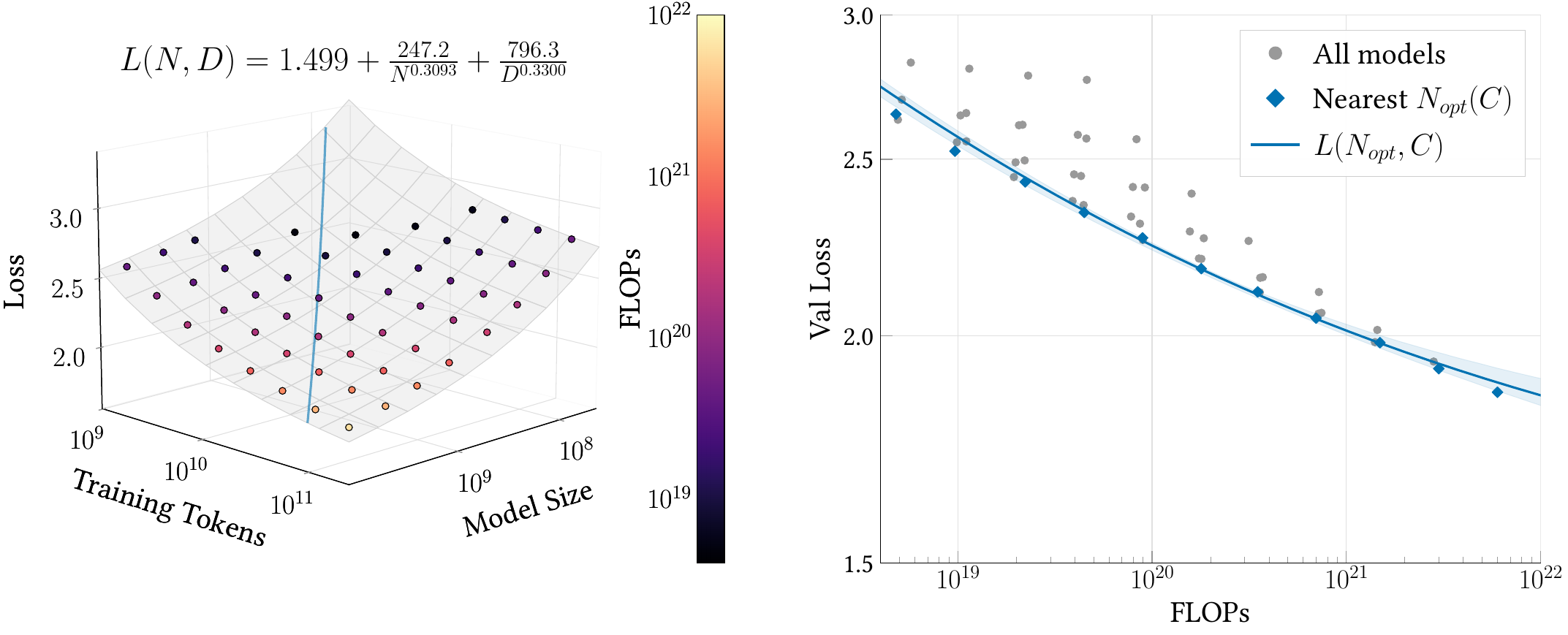}

\caption{\textbf{Validation loss scaling with model size and training tokens.} \textbf{Left:} Validation loss surface as a function of model size and training FLOPs for models trained on BETR Target-\coremetric (Nemotron-CC, top 10\% filtering). Contours show fitted scaling law values; the blue curve shows the compute-optimal model size $N_{\text{opt}}(C)$. \textbf{Right:} Validation loss along the compute-optimal path: gray dots show all trained models, blue highlights those closest to compute-optimal, and the line shows our fitted scaling curve $L(N_{\text{opt}}, C)$.}
\label{fig:app_nemotron_val_loss_scaling}
\end{figure}

Figure~\ref{fig:app_nemotron_val_loss_scaling} shows the fitted loss surface (Equation~\ref{eq:loss_scaling_law}) for one dataset. To identify compute-optimal models from this fit, we use the standard approximation $C = 6ND$ (FLOPs = 6 $\times$ parameters $\times$ tokens) to find $N_{opt}(C)$ (i.e. the model size that minimizes loss for each compute budget \citep{hoffmann2022training}). By binning the FLOPs range with logarithmic steps, we identify which trained models lie closest to this compute-optimal curve (Figure~\ref{fig:app_nemotron_val_loss_scaling}, left).

\begin{figure}[!h]
\centering
\includegraphics[width=\columnwidth]{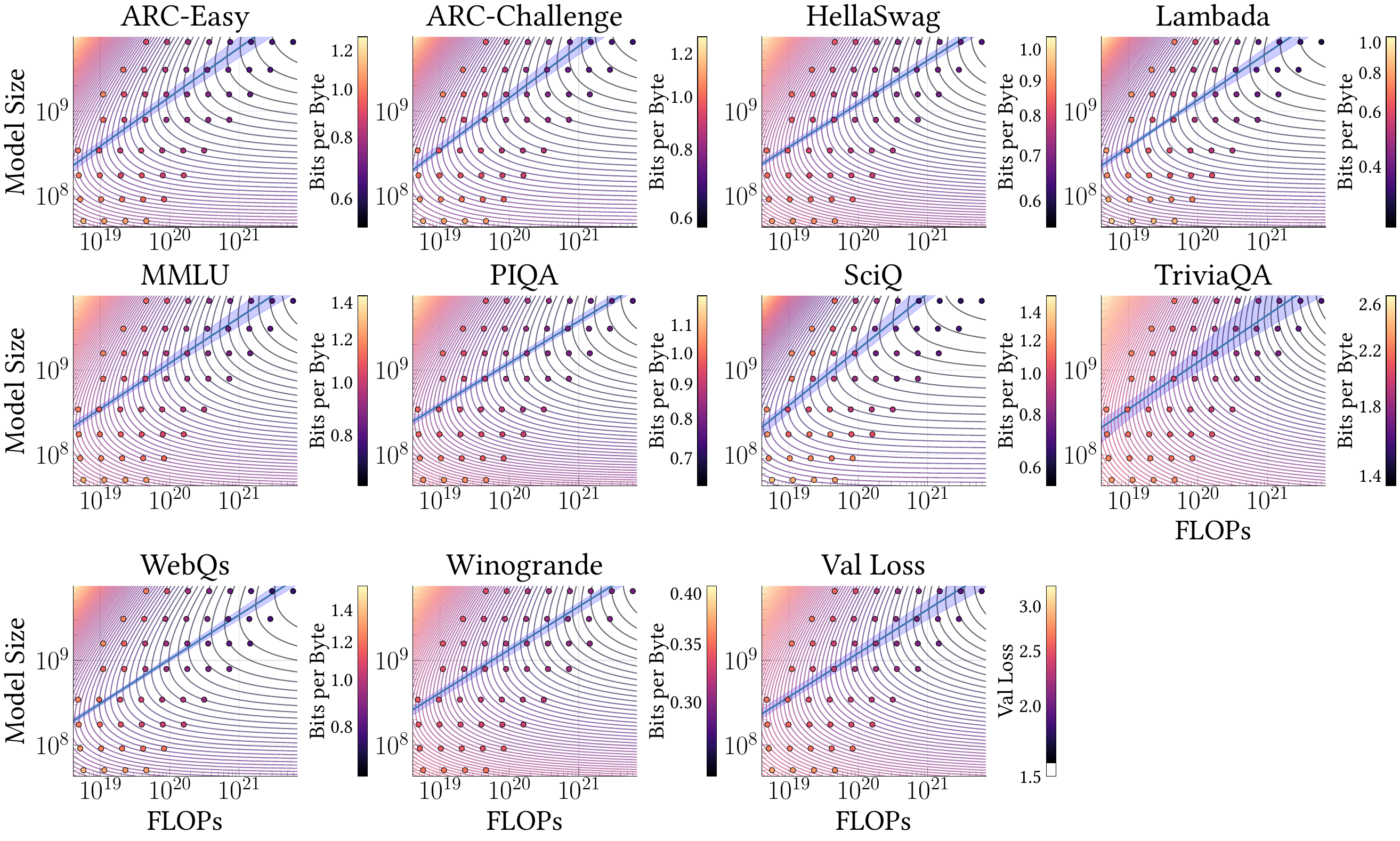}
\caption{\textbf{Task loss scaling with model size and training tokens.} Bits-per-byte loss surfaces for individual benchmarks and pretraining validation loss (bottom right) for BETR Target-\coremetric (Nemotron-CC, top 10\% filtering). Each panel shows how we fit scaling laws to every benchmark individually, with blue curves showing compute-optimal trajectories \(N_{\text{opt}}(C)\). These benchmark-specific fits enable the accuracy predictions shown in the main paper.}

\label{fig:app_nemotron_task_loss_scaling}
\end{figure}

Since our grid sampling creates higher density in mid-range FLOPs, we apply inverse density-based reweighting by bins in log-FLOPs space to ensure unbiased fits across all scales. Figure~\ref{fig:app_nemotron_task_loss_scaling} shows example fits for each benchmark, revealing that each has slightly different compute-optimal configurations. We therefore use validation loss to define compute-optimality for each dataset, as it provides a consistent reference across benchmarks. 

\begin{figure}[!h]
\centering
\includegraphics[width=\columnwidth]{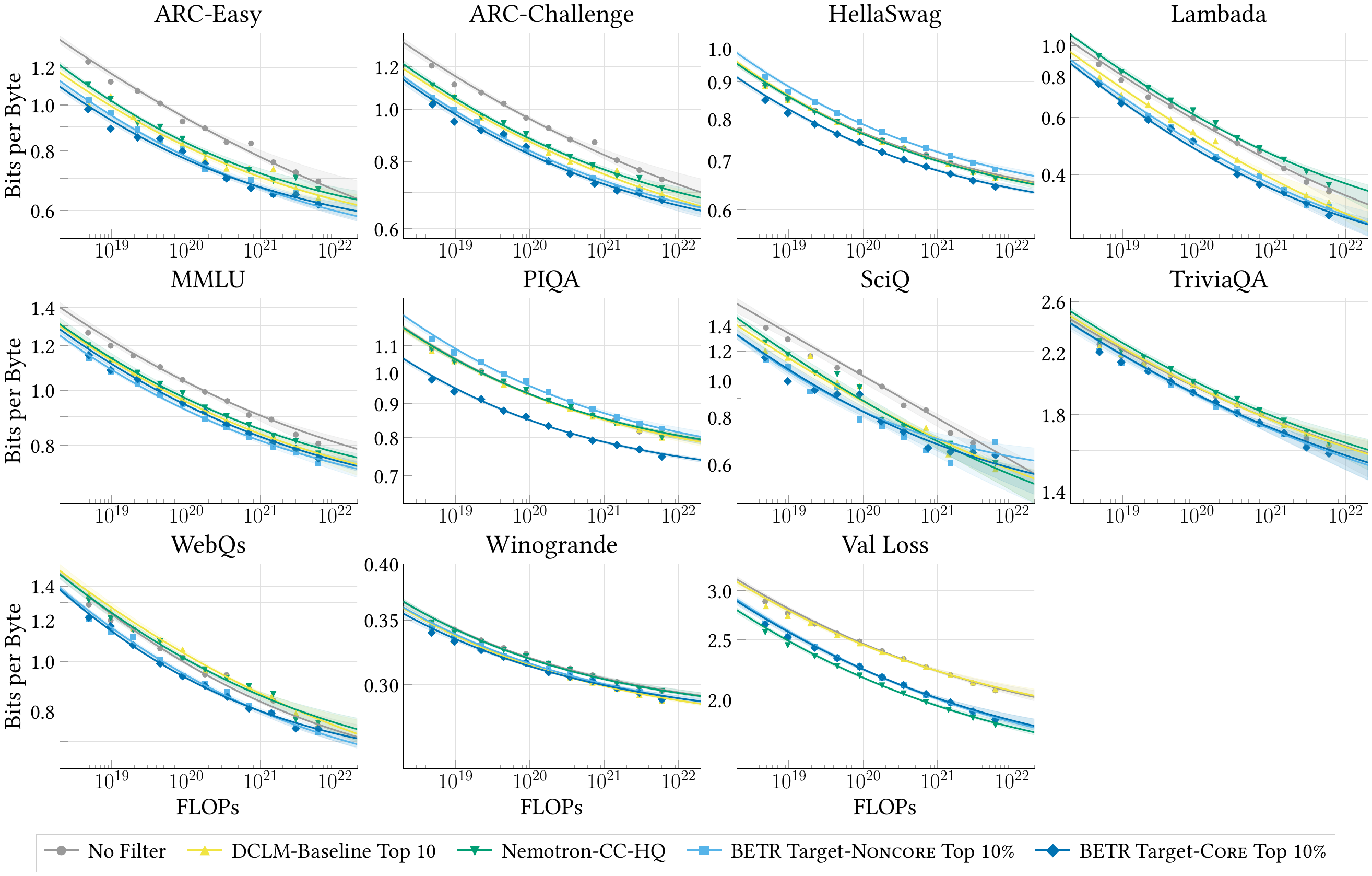}
\caption{\textbf{Task loss scaling comparison across data selection methods.} Compute-optimal performance on each benchmark for models trained on Nemotron-CC with different data selection methods. Lines show fitted scaling laws with 95\% confidence intervals (shaded). These per-benchmark fits show how each data selection method scales differently across tasks.}

\label{fig:app_nemotron_task_loss_scaling_opt_val_loss}
\end{figure}

We repeat this fitting procedure for validation loss and each benchmark's BPB across all datasets. Figure~\ref{fig:app_nemotron_task_loss_scaling_opt_val_loss} compares these fits across different data selection methods. Table~\ref{tab:app_loss_scaling_law_params} provides selected fitted parameters.

\FloatBarrier

\subsection{Task accuracy scaling}

We follow Llama 3's methodology~\citep{meta2024llama3} and model task accuracy by fitting each benchmark's accuracy as a shifted and scaled sigmoid function of its BPB:

\begin{equation}
\mathrm{Acc}_i(L_i) = \frac{c_1}{1 + e^{- k \cdot ( L_i - L_0 )}} + c_2.
\end{equation}

Here $\mathrm{Acc}_i$ is the accuracy of a model on benchmark $i$, $L_i$ is the BPB on the same benchmark. The constants $c_1$, $c_2$, $k$, and $L_0$ are fitting parameters.

For each benchmark task, we fit a separate sigmoid curve to the ($L_i$, $\text{Acc}_i$) pairs, minimizing the L2 loss between predicted and observed accuracy values. Unlike prior work that uses smoothing of consecutive intermediate checkpoints \citep{zhang2024does}, we only use accuracy and loss values for the final checkpoints of trained models. Following \cite{bhagia2024establishing}, we append a data point at $L=0.0$ and $Acc=1.0$ to each fit. We apply bootstrapping to estimate uncertainty, generating multiple fits by resampling the dataset (with replacement). This bootstrap distribution is propagated into our two-step prediction pipeline to estimate uncertainty in downstream predictions of task accuracy as a function of model size and training tokens.

\begin{figure}[!h]
\centering
\includegraphics[width=\columnwidth]{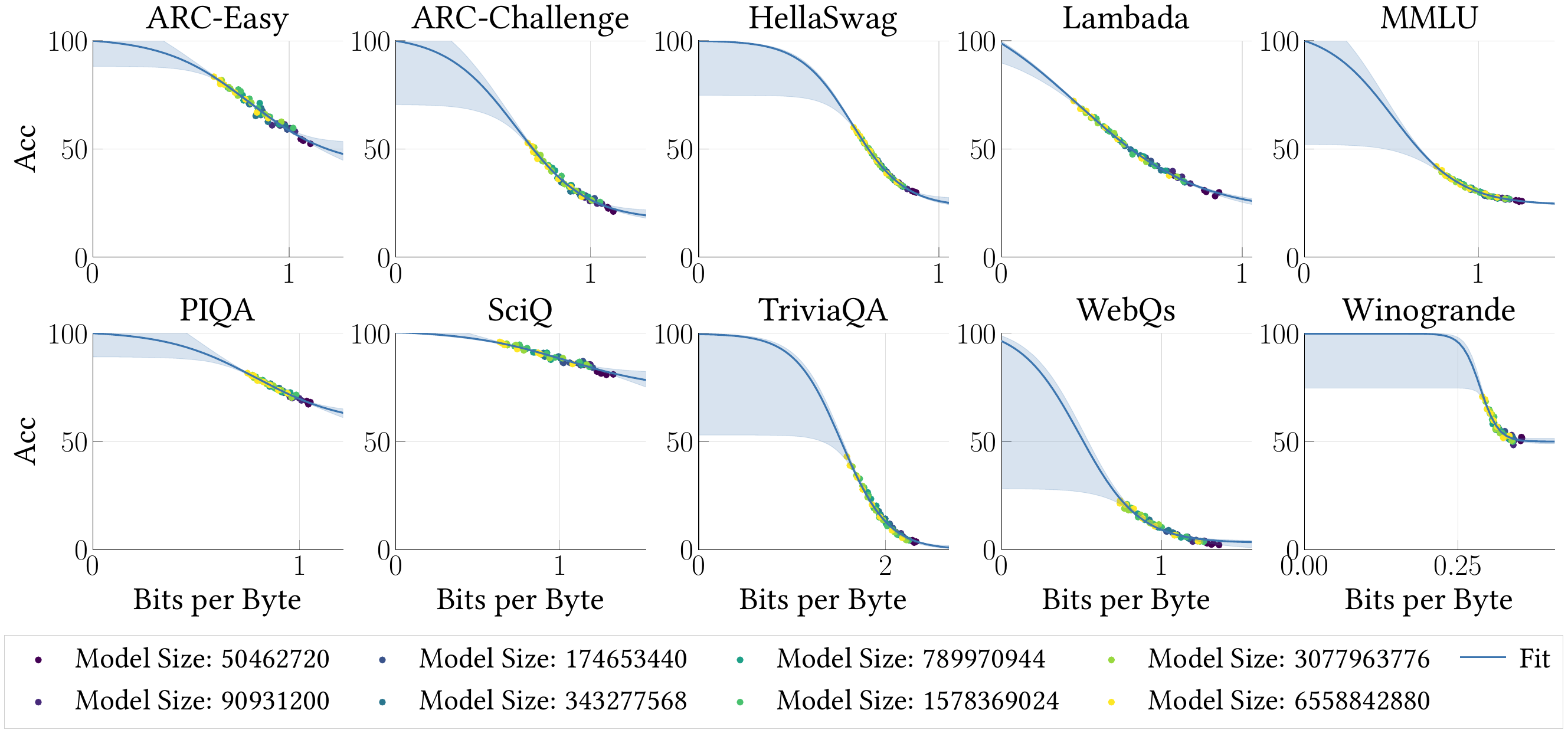}
\caption{\textbf{Task accuracy as a function of bits-per-byte.} Scaling law fits from the second stage of our two-step prediction pipeline, relating benchmark accuracy to task loss (in bits-per-byte) for models trained on the Nemotron-CC BETR Target-\coremetric dataset. Each subplot shows data points colored by model size and the corresponding sigmoid fit with 95\% bootstrap confidence intervals (blue band). These fits capture the saturating behavior of task accuracy as loss improves and are critical for translating loss scaling predictions into accuracy estimates.}
\label{fig:app_nemotron_comp_loss_to_acc}
\end{figure}

\begin{figure}[!h]
\centering
\includegraphics[width=\columnwidth]{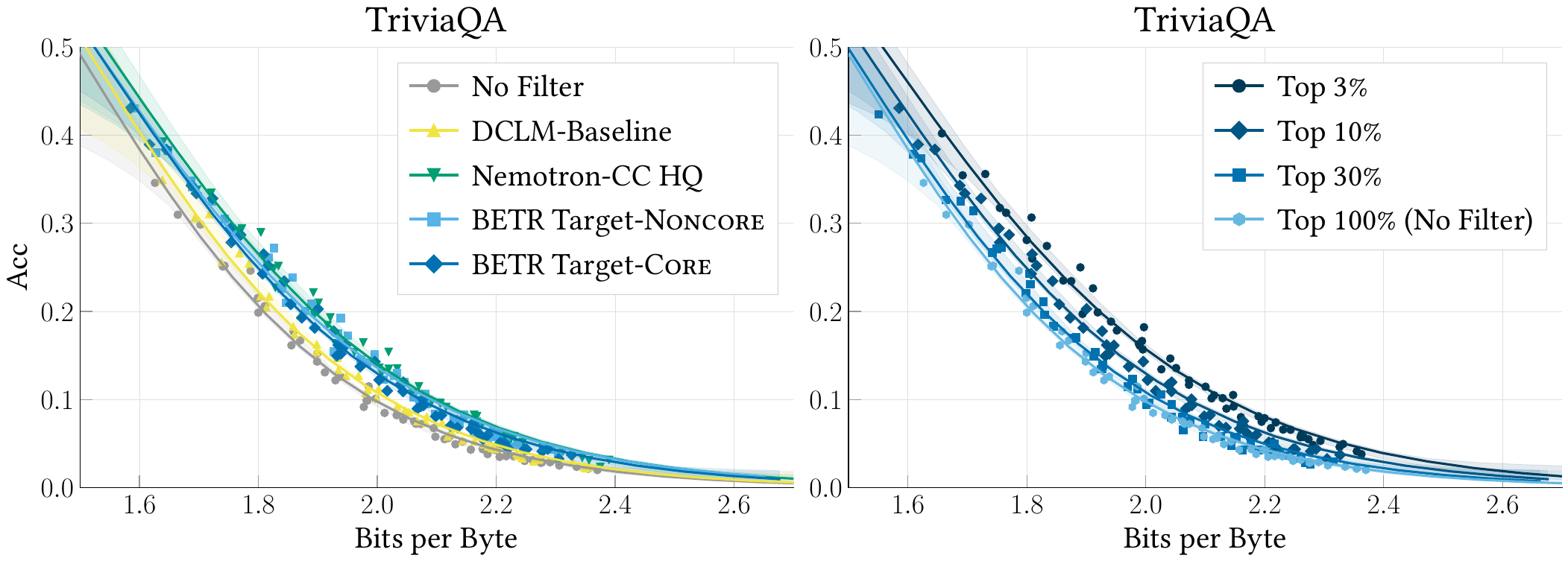}
\caption{\textbf{Impact of data selection methods and filtering intensity on accuracy-to-loss scaling.} Each plot shows the fitted relationship between TriviaQA bits-per-byte (BPB) and TriviaQA accuracy for models trained on Nemotron-CC with varying selection approaches. \textbf{Left:} Comparison across data selection methods. \textbf{Right:} Comparison across filtering intensities (Top 3\% to 100\%) for BETR Target-\coremetric. Bands denote 95\% bootstrap confidence intervals for the sigmoid fits.}
\label{fig:app_nemotron_triviaqa_loss_to_acc_comp}
\end{figure}

Importantly, we perform these fits independently per benchmark and per dataset, as we observe significant variations in the loss-to-accuracy relationship across different data pools, data selection methods, and filtering rates. Figure~\ref{fig:app_nemotron_comp_loss_to_acc} shows example accuracy-to-loss fits for the Nemotron-CC BETR Target-\coremetric dataset, while Figure~\ref{fig:app_nemotron_triviaqa_loss_to_acc_comp} illustrates how these relationships vary across different filtering strategies and intensities. Maintaining independent fits for each configuration is critical for accurate predictions.

Following \cite{bhagia2024establishing}, we combine our loss $L_i(N,D)$ and accuracy $\mathrm{Acc}_i(L_i)$ fits into two-step scaling law predictions of benchmark accuracy directly from model size and training tokens, as visualized in Figure~\ref{fig:nemotron_task_acc_scaling_comp}.

\FloatBarrier

\subsection{\coremetric average accuracy computation}

To compute the \coremetric average accuracy reported throughout the paper, we simply average the individual benchmark accuracies:
\begin{equation}
\mathrm{Acc}_{\mathrm{mean}}(N, D) = \frac{1}{|\mathcal{B}|} \sum_{i \in \mathcal{B}} \mathrm{Acc}_i(L_i(N,D)),
\end{equation}
where $\mathcal{B}$ denotes the set of \coremetric benchmarks. 

The bootstrap distributions from our two-step prediction pipeline naturally propagate through this averaging, providing confidence intervals for the mean accuracy at any compute budget. This enables us to determine which dataset outperforms another with quantified uncertainty (Figure~\ref{fig:pull_figure}) and to identify optimal filtering rates across scales (Figures~\ref{fig:optimal_filtering_scaling_main} and~\ref{fig:app_dclm_optimal_filtering_scaling}).

\FloatBarrier

\subsection{Compute multipliers}

Throughout the paper, we use compute multipliers (CMs) to compare data selection methods. A compute multiplier is the ratio of baseline compute to method compute needed for the same performance~\citep{betker_compute_2023,amodei2025deepseek}. Visually, this corresponds to a horizontal shift of the scaling curve (e.g., a 2x multiplier means the method's curve is shifted left by a factor of 2 on the compute axis).

To obtain robust estimates, we average this ratio across the full accuracy range. We:
\begin{enumerate}
\item Use fitted scaling laws to predict compute-optimal accuracy vs FLOPs for both methods
\item Discretize predictions into accuracy bins
\item Calculate mean FLOPs required per bin for each method
\item For overlapping bins, compute ratios (baseline FLOPs / method FLOPs)
\item Average these ratios
\end{enumerate}

Table~\ref{tab:app_efficiency_factors} reports the resulting compute multipliers, where values greater than 1 indicate the method outperforms the baseline (here, no filtering).

\input{tables/appendix/efficiency_factors}

\FloatBarrier

\section{Additional scaling law findings}
\label{app:additional_scaling_law_observations}

\subsection{Compute-optimal token-to-parameter ratios}

\begin{figure}[!h]
\centering
\includegraphics[width=\columnwidth]{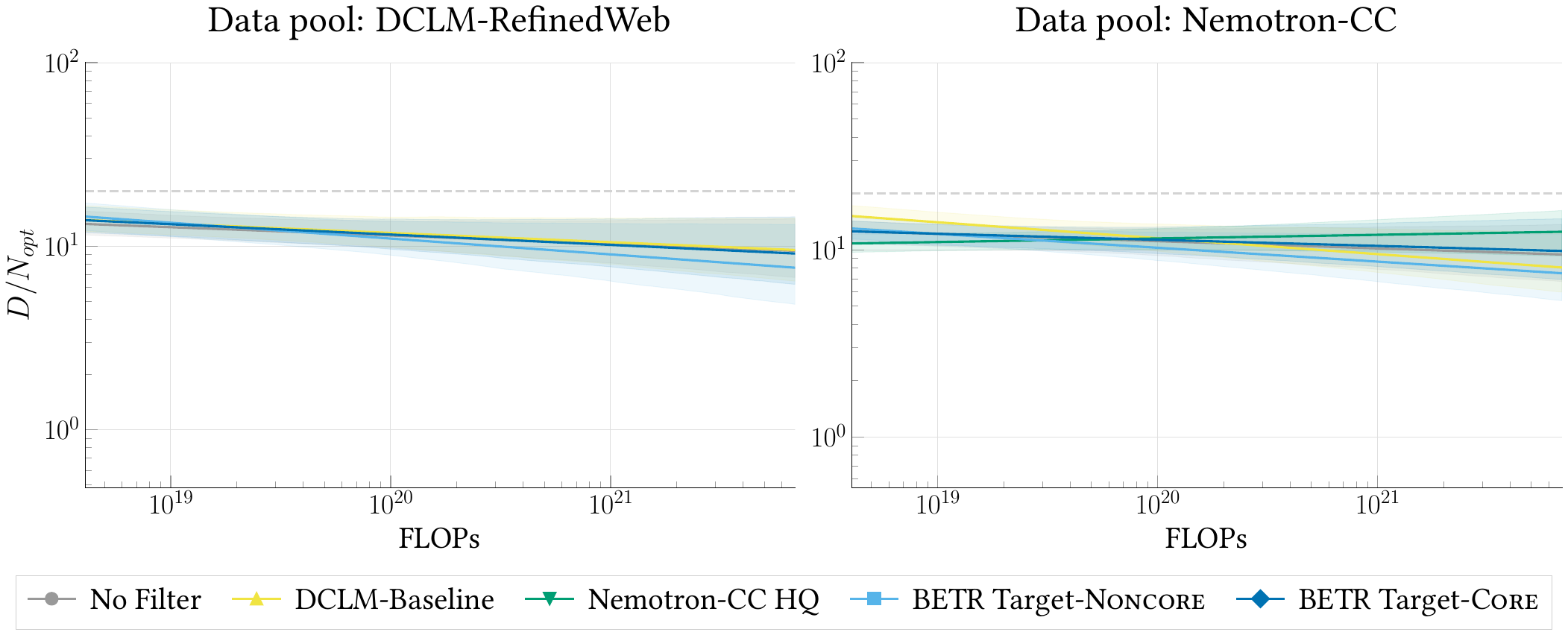}
\caption{\textbf{Optimal training token to model size ratio $D/N_{\text{opt}}$ across compute scales.} Ratio of training tokens $D$ to compute-optimal model size $N_{\text{opt}}(C)$ as a function of compute (FLOPs) for models trained on DCLM-RefinedWeb (left) and Nemotron-CC (right) data pools. Lines show the ratio for each data selection method, with shaded regions denoting 95\% confidence intervals from bootstrap resampling. The dashed horizontal line marks the Chinchilla ratio of $D/N \approx 20$~\citep{hoffmann2022training} All methods yield ratios around 10-12 tokens per parameter.}
\label{fig:app_d_over_n_opt}
\end{figure}

Figure \ref{fig:app_d_over_n_opt} shows the optimal training tokens to model size ratio $D/N_{\mathrm{opt}}$ as a function of compute across different data pools and selection methods. While \citet{hoffmann2022training} identified an optimal ratio of $\sim$20 tokens per parameter, our models consistently exhibit lower values, typically around 10-12 tokens per parameter.

This ratio remains stable: we observe no significant trends with increasing compute, and different data selection methods yield similar ratios within uncertainty bounds. This indicates that a constant token-to-parameter ratio may suffice across the compute scales we study, regardless of the data selection method used.

\FloatBarrier

\subsection{Per-benchmark scaling laws}

\begin{figure}[!h]
\centering
\includegraphics[width=\columnwidth]{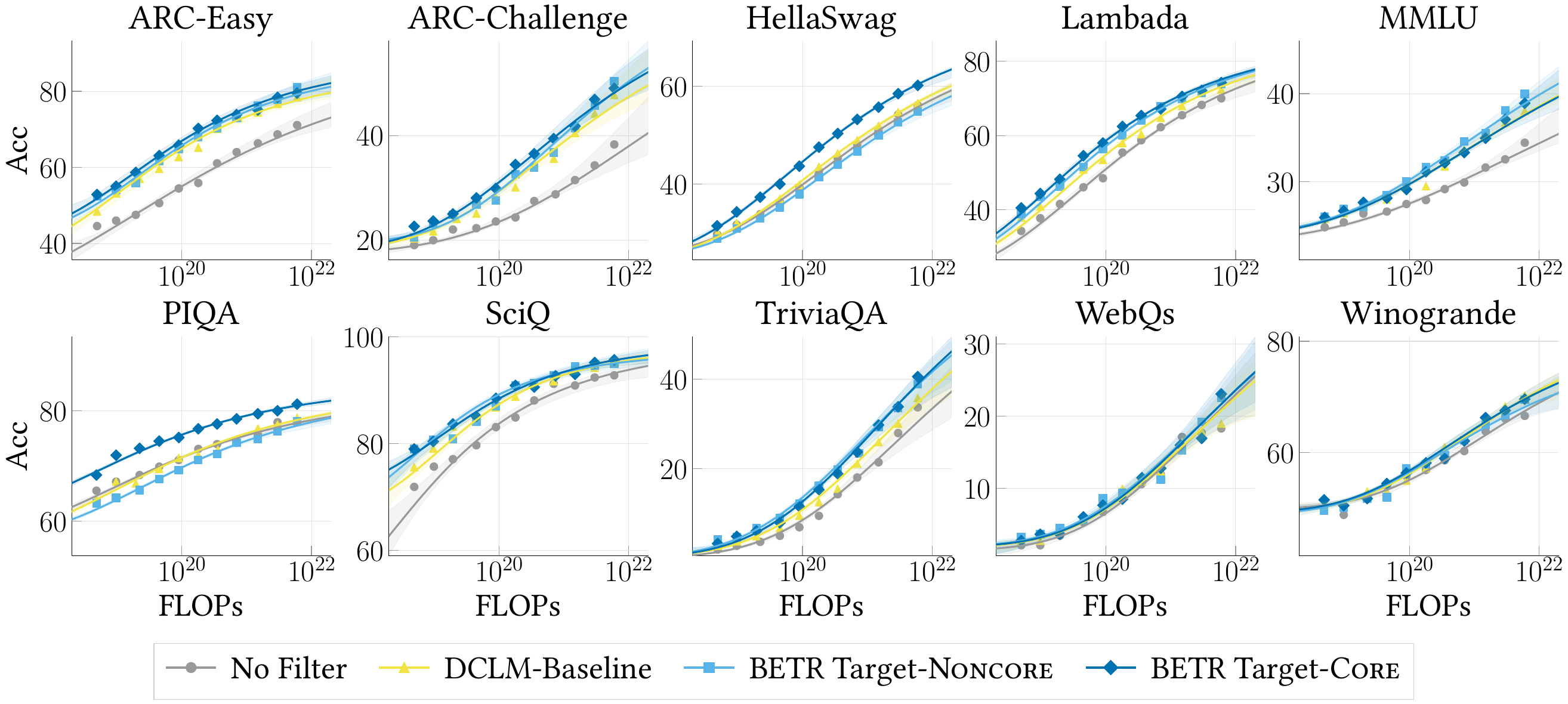}
\caption{\textbf{Per-benchmark accuracy scaling on DCLM-RefinedWeb.} Compute-optimal task accuracy for models trained on DCLM-RefinedWeb with different data selection methods.}
\label{fig:app_dclm_per_dataset_task_acc_scaling_comp}
\end{figure}

\begin{figure}[!h]
\centering
\includegraphics[width=\columnwidth]{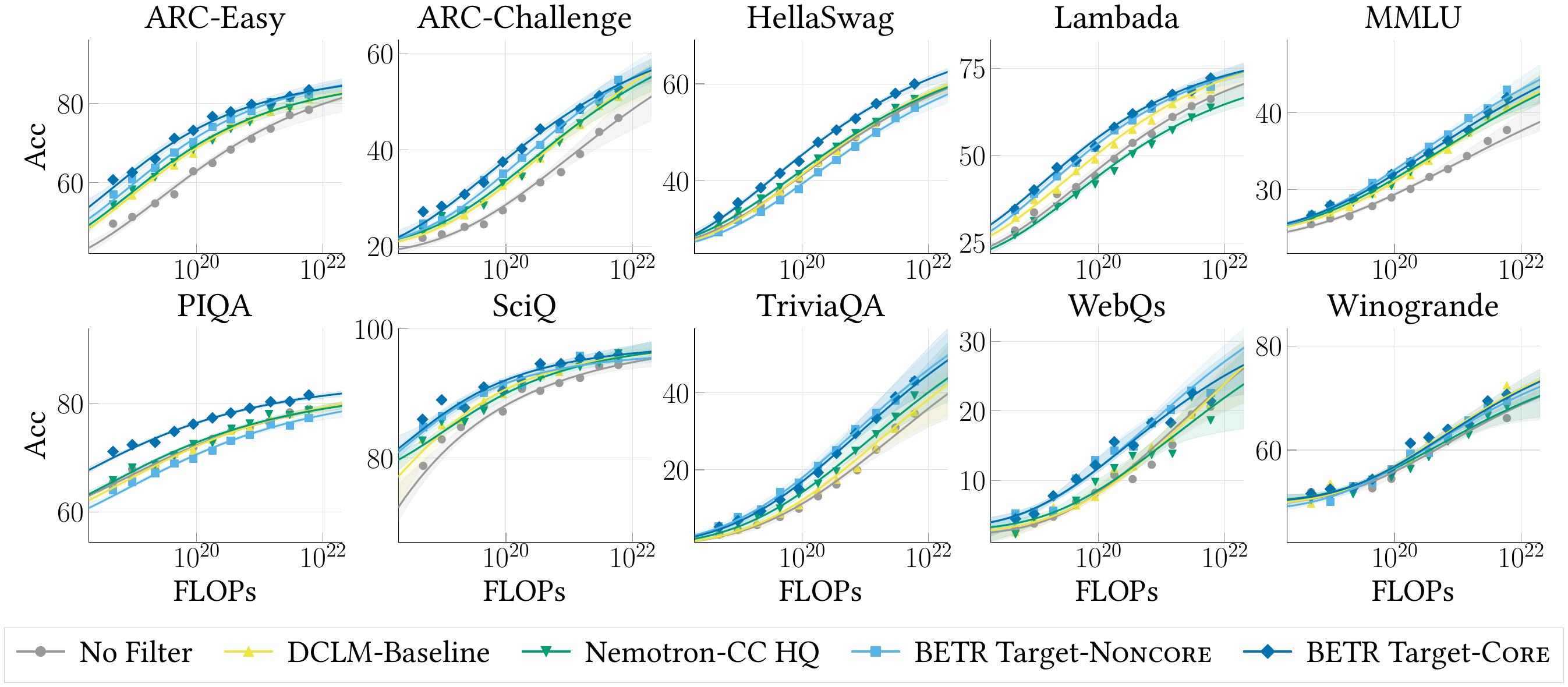}
\caption{\textbf{Per-benchmark accuracy scaling on Nemotron-CC.} Compute-optimal task accuracy for models trained on Nemotron-CC with different data selection methods.}
\label{fig:app_nemotron_per_dataset_task_acc_scaling_comp}
\end{figure}

Figures \ref{fig:app_dclm_per_dataset_task_acc_scaling_comp} and \ref{fig:app_nemotron_per_dataset_task_acc_scaling_comp} show per-benchmark accuracy scaling for compute-optimal models on DCLM-RefinedWeb and Nemotron-CC data pools, respectively. Across nearly all benchmarks, BETR Target-\coremetric consistently achieves the highest performance.

\subsection{Data pool comparison}

\begin{figure}[!h]
\centering
\includegraphics[width=0.7\textwidth]{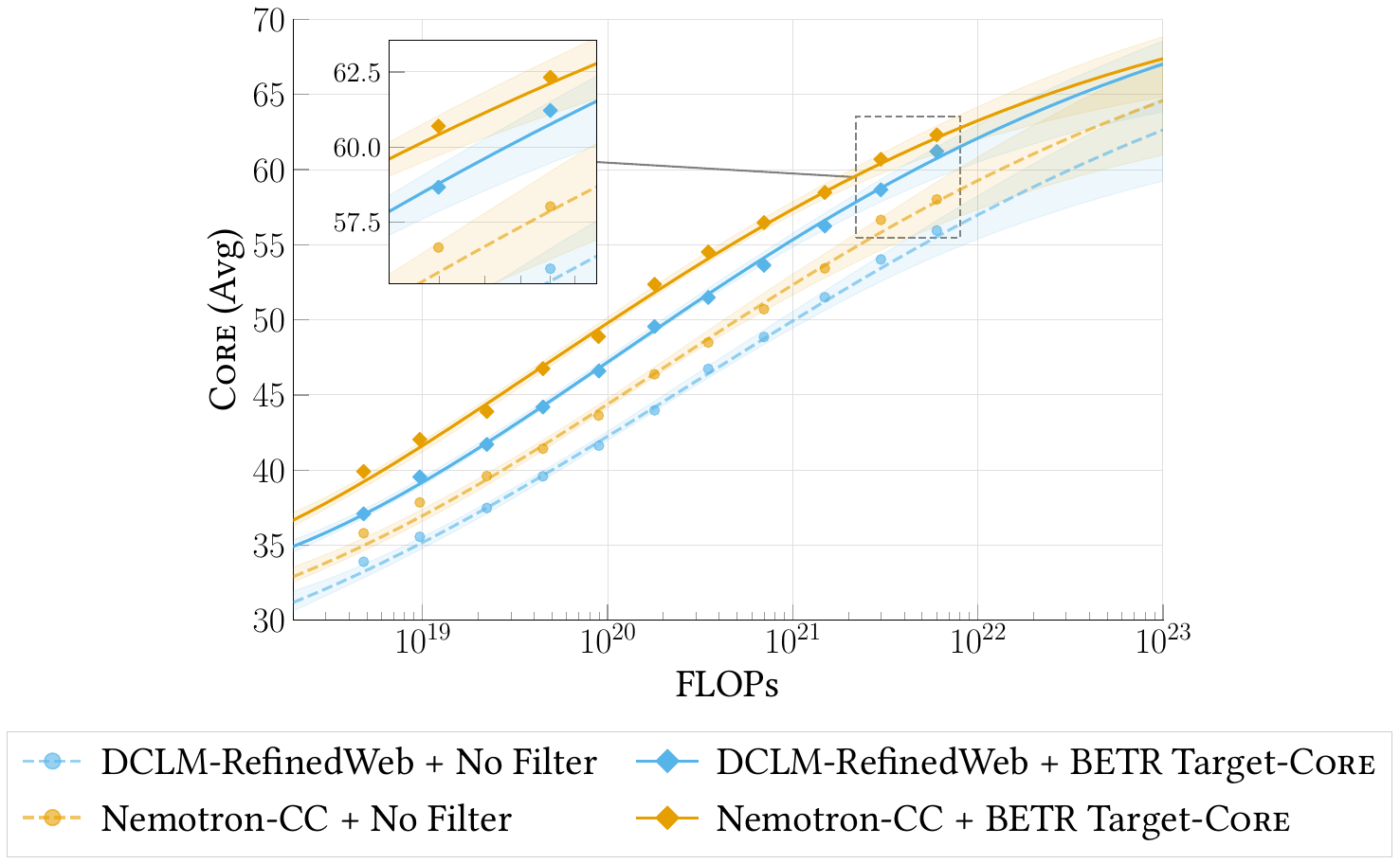}
\caption{\textbf{Data pool comparison with and without BETR filtering.} Mean benchmark accuracy (\coremetric average) as a function of compute for models trained on DCLM-RefinedWeb (\textcolor{wongOrange}{orange}) and Nemotron-CC (\textcolor{wongSky}{blue}), with (solid) and without (dashed) BETR Target-\coremetric filtering. Lines show scaling law predictions with 95\% confidence intervals; markers indicate actual trained models closest to compute-optimal.}
\label{fig:app_cross_datapool_comparison}
\end{figure}

Figure~\ref{fig:app_cross_datapool_comparison} directly compares \coremetric average accuracy across both data pools. Without filtering, Nemotron-CC consistently outperforms DCLM-RefinedWeb at all compute budgets. BETR Target-\coremetric filtering improves both data pools, with Nemotron-CC + BETR achieving the best overall performance. The scaling laws suggest the gap between data pools may narrow at higher compute budgets.

\subsection{Optimal filtering rates on DCLM-RefinedWeb}
\label{app:subsection_dclm_opt_filterin}

\begin{figure}[!ht]
\centering
\includegraphics[width=\columnwidth]{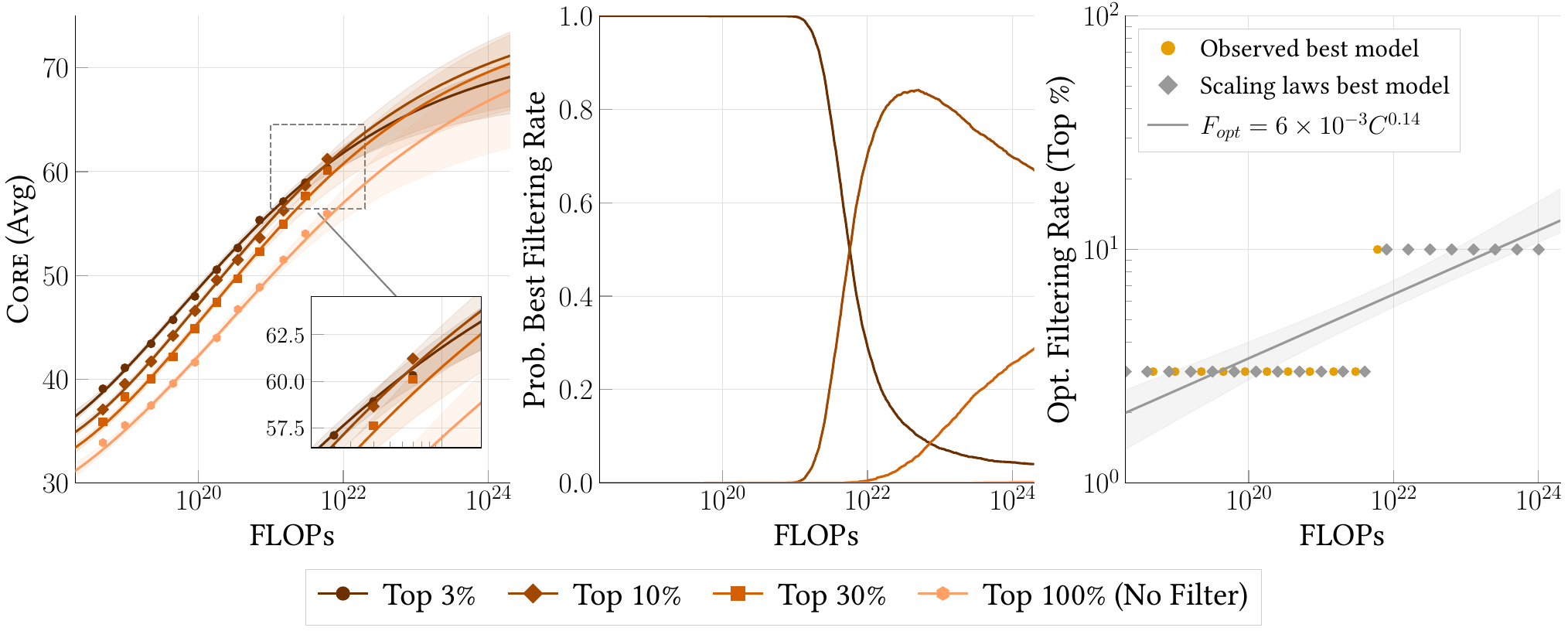}
\caption{\textbf{Optimal filtering rate scaling on DCLM-RefinedWeb.} \textbf{Left:} Mean task accuracy (\coremetric average) for models trained with different BETR Target-\coremetric filtering rates, with scaling law fits and 95\% confidence intervals. \textbf{Middle:} Probability that each filtering rate is optimal at a given compute scale, estimated from bootstrap distributions. \textbf{Right:} Optimal filtering rate (top-$x\%$) as a function of compute for observed models (\textcolor{wongOrange}{orange}) and scaling law predictions (\textcolor{gray}{gray}). The fitted relationship $F_{\mathrm{opt}}(C) = 6 \times 10^{-3} C^{0.14}$ is shown.}
\label{fig:app_dclm_optimal_filtering_scaling}
\end{figure}

Figure~\ref{fig:app_dclm_optimal_filtering_scaling} extends our filtering rate analysis to the DCLM-RefinedWeb data pool. As with Nemotron-CC (Figure~\ref{fig:optimal_filtering_scaling_main}), we observe a shift from aggressive filtering (Top 3\%) toward lighter filtering (Top 10\%) as compute increases. However, the transition is less pronounced: the optimal filtering rate scales as $F_{\text{opt}} \propto C^{0.14}$, compared to $C^{0.25}$ for Nemotron-CC. At $10^{23}$ FLOPs, the optimal filtering rate is predicted to be Top 10\%, compared to Top 30\% for Nemotron-CC at the same compute budget.

\FloatBarrier

\subsection{Fixed model size scaling}
\label{app:subsection_fixed_size}

\begin{figure}[htb]
  \centering
  \includegraphics[width=\columnwidth]{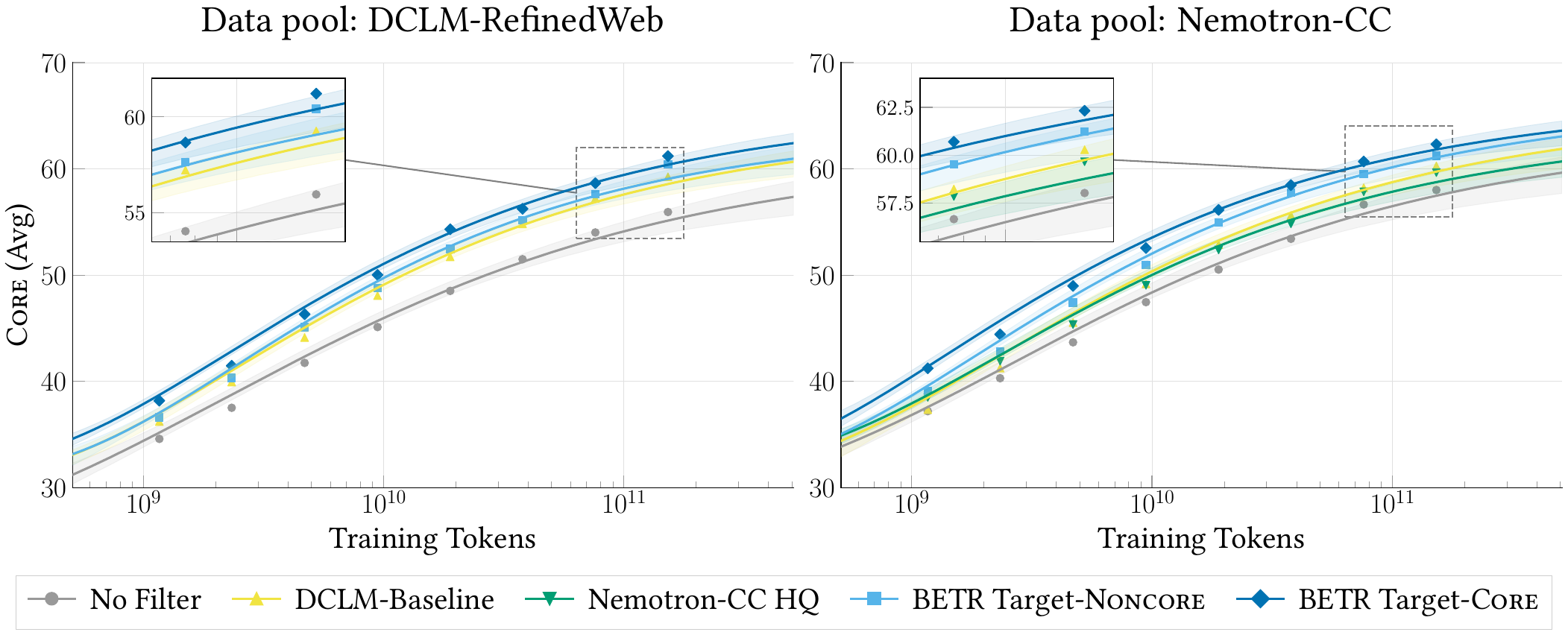}
    \caption{\textbf{Performance comparison at fixed model size.} Mean benchmark accuracy (\coremetric average) as a function of training tokens for 6.6B parameter models trained on DCLM-RefinedWeb (left) and Nemotron-CC (right) with various data selection methods. Lines show scaling law fits; markers indicate measured performance. BETR Target-\coremetric consistently outperforms baselines across all token counts. Inset plots magnify the high-token regime to highlight performance differences.}
  \label{fig:app_7b_comparison}
\end{figure}

Figure~\ref{fig:app_7b_comparison} compares mean benchmark accuracy across data selection methods for a fixed model size of 6.6B parameters as a function of training tokens. This analysis extends to the overtraining regime commonly used in production LLMs. The results show that the performance ranking observed under compute-optimal scaling persists at fixed model size: BETR Target-\coremetric consistently achieves the highest performance, outperforming DCLM-Baseline, Nemotron-CC HQ, and unfiltered data across all training budgets.

\begin{figure}[h]
\centering
\includegraphics[width=\columnwidth]{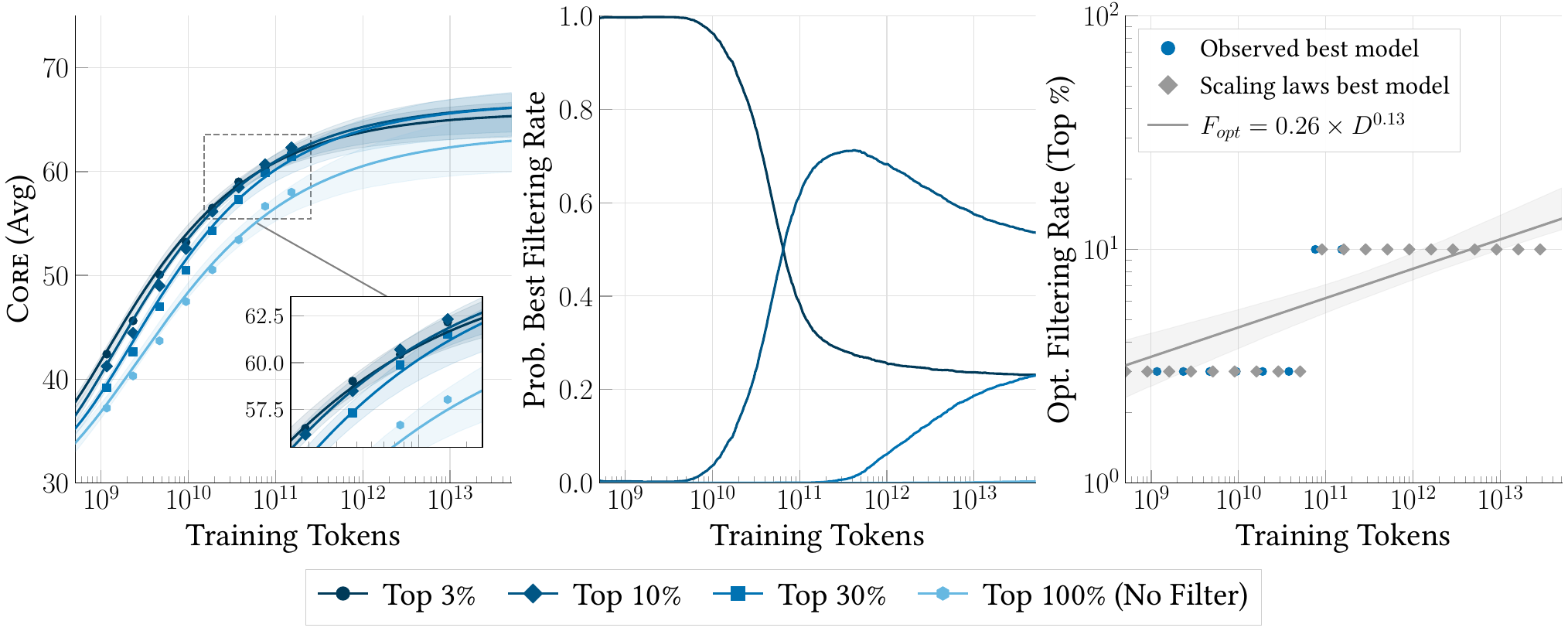}
\caption{\textbf{Optimal data filtering rate scaling at fixed model size.} \textbf{Left:} Mean task accuracy (\coremetric average) for 6.6B models trained with different BETR Target-\coremetric filtering rates on Nemotron-CC, along with scaling law fits. \textbf{Center:} Probability that each filtering rate is optimal at a given number of training tokens, estimated from bootstrap distributions of scaling law fits. \textbf{Right:} Optimal filtering rate (top-$x\%$) as a function of training tokens, shown both for the \textcolor{wongBlue}{best observed model} and the \textcolor{gray}{best model predicted via scaling laws}. The scaling law fit $F_{\mathrm{opt}}(D) = 0.26 \times D^{0.13}$ is overlaid for comparison.}
\label{fig:app_nemotron_cc_optimal_filtering_scaling_at_6p6b}
\end{figure}

Figure~\ref{fig:app_nemotron_cc_optimal_filtering_scaling_at_6p6b} investigates how optimal filtering rates change with training tokens at fixed 6.6B model size. Unlike compute-optimal scaling, we do not observe a transition from 10\% to 30\% filtering within the sub-$10^{24}$ FLOPs range. However, a transition from 3\% to 10\% filtering still occurs as training progresses. The optimal filtering rate scales as $F_{\text{opt}}(D) \propto D^{0.13}$, slower than the $F_{\text{opt}}(C) \propto C^{0.25}$ observed under compute-optimal conditions. This shows that optimal filtering strategies differ between fixed model size scaling and compute-optimal scaling.

\FloatBarrier

\subsection{Example scaling law parameters}

Tables~\ref{tab:app_loss_scaling_law_params} and~\ref{tab:app_acc_scaling_law_params} present example scaling law parameters for models trained on Nemotron-CC with BETR Target-\coremetric filtering at various filtering rates. For brevity, we show only a representative subset of the complete parameter fits. Each parameter includes 95\% confidence intervals derived from bootstrap distributions. Note that these parameter-level uncertainties do not directly translate to the prediction uncertainties shown in our plots, which are computed from the full ensemble of bootstrap models.

%% file: tables/appendix/efficiency_factors.tex
\begin{table}[ht]
\centering
\begin{adjustbox}{width=\textwidth}
{%
  \setlength{\tabcolsep}{0.5pt}      %
\renewcommand{\arraystretch}{1.0}  %
  \begin{tabular}{
    l@{\hspace{8pt}}    %
    *{10}{G}            %
    @{\hspace{5pt}}c   %
  }
  \toprule
  Method & ARC-E & ARC-C & HellaS & Lamb. & MMLU & PIQA & SciQ & TrQA & WebQs & Wino. & \coremetricavg \\
  \midrule
  \multicolumn{12}{l}{\textbf{Data pool: DCLM-RefinedWeb}} \\ \midrule
  No Filter                          & 1.0 & 1.0 & 1.0 & 1.0 & 1.0 & 1.0 & 1.0 & 1.0 & 1.0 & 1.0 & 1.0 \\
  DCLM-Baseline   & 10.2 & 8.7  & 1.2 & 1.9 & 5.7 & 1.0 & 3.2 & 1.9 & 1.1 & 1.7 &  2.6 \\
  BETR Target-\extendedmetric        & 13.5 & 10.2 & 0.7 & 2.8 & \textbf{7.9} & 0.5 & \textbf{5.7} & \textbf{3.4} & \textbf{1.4} & 1.5 & 3.1 \\
  \highlight[ontarget]{BETR Target-\coremetric} & \textbf{17.9} & \textbf{13.8} & \textbf{3.1} & \textbf{3.6} & 5.7 & \textbf{9.7} & 5.3 & 3.0 & 1.3 & \textbf{1.9} & \textbf{4.7} \\
  \midrule
  \multicolumn{12}{l}{\textbf{Data pool: Nemotron-CC}} \\ \midrule
  No Filter                                     & 1.0 & 1.0 & 1.0 & 1.0 & 1.0 & 1.0 & 1.0 & 1.0 & 1.0 & 1.0 & 1.0 \\ 
  Nemotron-CC HQ                                & 3.7 & 3.5 & 1.3 & 0.7 & 3.8 & 1.2 & 2.7 & 2.0 & 1.0 & 1.3 & 1.7 \\
  DCLM-Baseline              & 3.2 & 3.1 & 1.1 & 2.2 & 3.3 & 0.9 & 2.9 & 1.3 & 1.0 & \textbf{1.9} & 1.9 \\
  BETR Target-\extendedmetric                   & 6.1 & 5.2 & 0.6 & 3.3 & \textbf{6.6} & 0.4 & 5.0 & \textbf{4.4} & \textbf{3.2} & 1.5 & 3.0 \\
  \highlight[ontarget]{BETR Target-\coremetric} & \textbf{10.4} & \textbf{8.2} & \textbf{2.7} & \textbf{4.5} & 5.2 & \textbf{9.5} & \textbf{6.9} & 3.6 & 2.8 & 1.8 & \textbf{4.7} \\
  \bottomrule
  \end{tabular}
}%
\end{adjustbox}
\caption{\textbf{Compute multipliers for BETR and baselines.} Compute multipliers relative to no filtering baseline for each data pool. BETR Target-\coremetric \highlight[ontarget]{directly optimizes} for the evaluation benchmarks and achieves the highest average compute multiplier. \textcolor{tabgray}{\textbf{Bold}: Highest compute multiplier per benchmark for each data pool.}}
\label{tab:app_efficiency_factors}
\end{table}

%% file: sec/8_app_extra_results.tex
\section{Additional experiments \& analysis}

\subsection{Detailed \extendedmetric results}
To complement Figure~\ref{fig:core_vs_noncore_performance}, we report the full \extendedmetric results at 7B-10x scale in Table ~\ref{tab:app_noncore_results_10x}. The benchmarks in each subcategory are listed in Table~\ref{tab:app_noncore_benchmark_list}.

\input{tables/appendix/noncore_results_7b_10x}

\FloatBarrier

\subsection{7B-1x \coremetric results}

For completeness, we also report \coremetric performance at 7B-1x scale for BETR variants and baselines in Table~\ref{tab:app_main_results_7b_1x}.

\input{tables/appendix/main_results_7b_1x}

\subsection{Nemotron-CC HQ dataset definition}
\label{app:nemotron_cc_hq_diff}

Throughout this work, we use Nemotron-CC HQ to refer to all documents labeled as ``high-quality'' from the Nemotron-CC data pool, which includes both real documents and all types of synthetic rewrites. This slightly differs from the dataset used by \citet{su2024nemotron}, which selects only high-quality real documents and high-quality synthetic documents specifically from the diverse QA pairs rewriting style, excluding other synthetic rewriting styles.

To understand the impact of this difference, we compare these two versions along with other Nemotron-CC subsets at 7B-1x scale in Table~\ref{tab:app_nemotron_cc_subset_comparison}. While \citet{su2024nemotron}'s more selective version performs slightly better ($\sim$1 std better) than our broader definition, this improvement is not enough to change method rankings.

\input{tables/appendix/nemotron_cc_comparison}

\FloatBarrier

\subsection{Per-benchmark standard deviations}

To estimate the natural variation in benchmark performance, we trained 10 models with different random initialization and data shuffling on a representative dataset (BETR Target-\coremetric on DCLM-RefinedWeb at 7B-1x scale). Table~\ref{tab:app_benchmark_stddev} reports the standard deviations across these runs.

\input{tables/appendix/bench_stddev}

\FloatBarrier

\clearpage

\subsection{Benchmark contributions to document selection}

When using BETR with max aggregation, each document in the top 10\% is selected because one benchmark example ranks it highly. Table~\ref{tab:app_benchmark_contribution} shows which benchmarks' examples are responsible for selecting documents when targeting all \coremetric benchmarks jointly and equally (using the same number of target examples per benchmark). The distribution forms two clusters: six benchmarks (TriviaQA, HellaSwag, MMLU, WebQs, WinoGrande, PIQA) each contribute 10--15\% of selected documents, while four benchmarks (Lambada, SciQ, ARC-Easy/Challenge) contribute 5--7\%. While not uniform, the distribution remains reasonably balanced with all benchmarks contributing meaningful fractions and no single benchmark dominating. This pattern holds consistently across both DCLM-RefinedWeb and Nemotron-CC data pools.

\input{tables/appendix/benchmark_contribution}

\FloatBarrier

\clearpage

\subsection{Topic and format distributions}

To show the content composition of different data selection methods, we compute topic and format distributions using WebOrganizer~\citep{wettig2025organize} models. These are computed on a subsample of 1M documents from DCLM-RefinedWeb filtered to the top 10\% of tokens. Figure~\ref{fig:app_topic_format_distribution} shows these distributions, providing a high-level view of content differences between methods.

\input{figures/appendix/topic_format_distribution}

\FloatBarrier

\section{Full tables}

The following tables present the full list of \extendedmetric benchmarks and example scaling law parameters from our experiments.

\input{tables/appendix/noncore_benchmark_list}

\input{tables/appendix/loss_scaling_law_parameters}

\input{tables/appendix/acc_scaling_law_parameters}

%% file: tables/appendix/noncore_results_7b_10x.tex
\begin{table}[htpb]
\centering
\begin{adjustbox}{max width=\textwidth}
{%
  \setlength{\tabcolsep}{2pt}      %
\renewcommand{\arraystretch}{1.0}  %
  \begin{tabular}{
    l@{\hspace{5pt}}    %
    c                   %
    @{\hspace{5pt}}     %
    *{5}{>{\centering\arraybackslash}p{5.4em}}      %
  }
  \toprule
  & & \multicolumn{5}{c}{\extendedmetric Subcategories} \\
  \cmidrule(lr){3-7}
  Method & \extendedmetric (Avg) & 
  \centering \footnotesize{Commonsense Reasoning} & 
  \centering \footnotesize{Language Understanding} & 
  \centering \footnotesize{Reading Comprehension} & 
  \centering \footnotesize{Symbolic Problem Solving} & 
  \centering\arraybackslash \footnotesize{World \ \ \ \ Knowledge} \\
  \midrule
  \multicolumn{7}{l}{\textbf{Data pool: DCLM-RefinedWeb}} \\ \midrule
  No Filter                          & 44.2 & 69.1 & 44.2 & 62.5 & 22.5 & 34.7 \\
  DCLM-Baseline   & 46.3 & \textbf{71.4} &\textbf{ 49.8} & 65.2 & 23.4 & 36.6 \\
  \highlight[ontarget]{BETR Target-\extendedmetric}        & \textbf{46.4} & 71.2 & 47.7 & \textbf{65.9} & \textbf{23.6} & \textbf{37.6} \\
  BETR Target-\coremetric & 44.2 & 69.8 & 47.7 & 62.8 & 20.8 & 35.8 \\
  \midrule
  \multicolumn{7}{l}{\textbf{Data pool: Nemotron-CC}} \\ \midrule
  No Filter                          & 45.0 & 70.8 & 47.2 & 62.2 & 22.7 & 37.3 \\ 
  Nemotron-CC HQ                     & 47.7 & 72.8 & 50.4 & 68.2 & 23.7 & 39.9 \\
  DCLM-Baseline  & 50.4 & 73.4 & 53.6 & 70.0 & \textbf{28.7} & 40.0 \\
  \highlight[ontarget]{BETR Target-\extendedmetric}        & \textbf{50.6} & \textbf{74.1} & \textbf{54.1} & \textbf{70.6} & 28.1 & \textbf{40.3} \\
  BETR Target-\coremetric & 48.5 & 73.9 & 50.9 & 67.9 & 25.5 & 38.5 \\
  \bottomrule
  \end{tabular}
}%
\end{adjustbox}
\begin{flushleft}
\caption{\textbf{\extendedmetric results at 7B-10x scale.} We compare BETR variants and baselines at 7B-10x scale (7B parameters, 1.4T tokens) on \extendedmetric tasks. BETR Target-\extendedmetric \highlight[ontarget]{directly optimizes} for these benchmarks and achieves the highest average performance, followed closely by DCLM-Baseline. \textcolor{tabgray}{\textbf{Bold}: best result.}}
\label{tab:app_noncore_results_10x}
\end{flushleft}
\end{table}

%% file: tables/appendix/main_results_7b_1x.tex
\begin{table}[htpb]
\centering
\begin{adjustbox}{width=\textwidth}
{%
  \setlength{\tabcolsep}{0.5pt}      %
\renewcommand{\arraystretch}{1.0}  %
  \begin{tabular}{
    l@{\hspace{8pt}}    %
    *{10}{G}            %
    @{\hspace{5pt}}c   %
  }
  \toprule
  Method & ARC-E & ARC-C & HellaS & Lamb. & MMLU & PIQA & SciQ & TrQA & WebQs & Wino. & \coremetricavg \\
  \midrule
  \multicolumn{12}{l}{\textbf{Data pool: DCLM-RefinedWeb}} \\ \midrule
  No Filter                          & 72.3 & 39.0 & 56.1 & 71.2 & 27.4 & 78.2 & 94.0 & 33.7 & 19.8 & 68.4 & 56.0 \\
  DCLM-Baseline   & \textbf{79.4} & 48.1 & 57.0 & 72.7 & \textbf{52.8} & 77.7 & \textbf{95.2} & 35.7 & 20.1 & \textbf{71.4} & 61.0 \\
  BETR Target-\extendedmetric & \textbf{79.2} & 48.0 & 55.0 & 73.0 & \textbf{53.3} & 77.0 & 94.7 & 38.3 & \textbf{22.2} & 68.4 & 60.9 \\
  \highlight[ontarget]{BETR Target-\coremetric}        & \textbf{79.7} & \textbf{49.7} & \textbf{59.9} & \textbf{74.0} & 48.9 & \textbf{81.3} & \textbf{95.4} & \textbf{40.4} & \textbf{22.5} & \textbf{71.0} & \textbf{62.3} \\
  \midrule
  \multicolumn{12}{l}{\textbf{Data pool: Nemotron-CC}} \\ \midrule
  No Filter                          & 78.8 & 44.9 & 56.3 & 65.9 & 50.7 & 78.2 & 95.1 & 36.3 & 21.1 & 67.9 & 59.5 \\ 
  Nemotron-CC HQ                     & 81.0 & 51.6 & 56.8 & 63.6 & 58.0 & 79.4 & 95.2 & 36.5 & 17.5 & 68.2 & 60.8 \\
  DCLM-Baseline  & 79.9 & 51.9 & 57.2 & 69.6 & 56.7 & 78.9 & \textbf{96.5} & 36.1 & 22.2 & \textbf{72.7} & 62.2 \\
  BETR Target-\extendedmetric & \textbf{83.0} & \textbf{54.8} & 55.2 & 70.0 & \textbf{59.9} & 77.8 & \textbf{96.2} & \textbf{43.1} & 22.1 & 69.0 & 63.1 \\
  \highlight[ontarget]{BETR Target-\coremetric}                 & \textbf{83.4} & \textbf{54.1} & \textbf{59.9} & \textbf{71.5} & \textbf{59.4} & \textbf{81.6} & 95.8 & \textbf{43.0} & \textbf{25.3} & \textbf{72.2} & \textbf{64.6} \\
  \bottomrule
  \end{tabular}
}%
\end{adjustbox}
\caption{\textbf{\coremetric results at 7B-1x scale.} Comparison of BETR variants and baselines at 7B-1x scale (7B parameters, 140B tokens). BETR Target-\coremetric \highlight[ontarget]{directly optimizes} for these benchmarks and achieves the highest average performance. \textcolor{tabgray}{\textbf{Bold}: best result $\pm$ 1 std.}}
\label{tab:app_main_results_7b_1x}
\end{table}

%% file: tables/appendix/nemotron_cc_comparison.tex
\begin{table}[htpb]
\centering
\begin{adjustbox}{width=\textwidth}
{%
  \setlength{\tabcolsep}{0.5pt}      %
\renewcommand{\arraystretch}{1.0}  %
  \begin{tabular}{
    l@{\hspace{8pt}}    %
    *{10}{G}            %
    @{\hspace{5pt}}c   %
  }
  \toprule
  Nemotron-CC Subset & ARC-E & ARC-C & HellaS & Lamb. & MMLU & PIQA & SciQ & TrQA & WebQs & Wino. & \coremetricavg \\
  \midrule
  All                                & 78.8 & 44.9 & 56.3 & 65.9 & 50.7 & 78.2 & 95.1 & 36.3 & 21.1 & 67.9 & 59.5 \\ 
  Actual Only                        & 75.3 & 40.7 & 56.3 & 67.2 & 28.8 & 77.8 & 93.9 & 32.5 & 16.7 & 67.1 & 55.6 \\
  Synthetic Only                     & 80.5 & 49.6 & 55.9 & 62.8 & 55.4 & 78.8 & 95.7 & 38.7 & 21.5 & 67.6 & 60.6 \\
  HQ (Actual + Synthetic QA)$^\dagger$ & 81.5 & 51.0 & 57.2 & 64.2 & 58.6 & 78.5 & 95.4 & 37.3 & 19.9 & 69.5 & 61.3 \\
  HQ (All)$^\ddagger$                & 81.0 & 51.6 & 56.8 & 63.6 & 58.0 & 79.4 & 95.2 & 36.5 & 17.5 & 68.2 & 60.8 \\
  \bottomrule
  \end{tabular}
}%
\end{adjustbox}
\caption{\textbf{Performance of different Nemotron-CC data subsets} at 7B-1x scale (7B parameters, 140B tokens). $^\dagger$HQ (Actual + Synthetic Diverse QA) corresponds to the subset used in \citet{su2024nemotron}. $^\ddagger$HQ (All) is the definition used throughout this paper as Nemotron-CC HQ.}
\label{tab:app_nemotron_cc_subset_comparison}
\end{table}

%% file: tables/appendix/bench_stddev.tex
\begin{table}[h]
\centering
\begin{adjustbox}{width=\textwidth}
{%
  \setlength{\tabcolsep}{0.5pt}      %
  \renewcommand{\arraystretch}{1.0}  %
  \begin{tabular}{
    l@{\hspace{8pt}}    %
    *{10}{G}            %
    @{\hspace{5pt}}c    %
  }
  \toprule
  & ARC-E & ARC-C & HellaS & Lamb. & MMLU & PIQA & SciQ & TrQA & WebQs & Wino. & \coremetricavg \\
  \midrule
  Std. Dev. & 0.49 & 0.85 & 0.18 & 0.48 & 1.10 & 0.44 & 0.32 & 0.84 & 1.33 & 1.15 & 0.24 \\
  \bottomrule
  \end{tabular}
}%
\end{adjustbox}
\caption{\textbf{Standard deviations of benchmark performance} across 10 runs with different random initialization and data shuffling, using BETR Target-\coremetric on DCLM-RefinedWeb at 7B-1x scale (7B parameters, 140B tokens).}
\label{tab:app_benchmark_stddev}
\end{table}

%% file: tables/appendix/benchmark_contribution.tex
\begin{table}[htpb]
\centering

\begin{tabular}{lcc}
\toprule
& \multicolumn{2}{c}{Contribution (\%)} \\
\cmidrule(lr){2-3}
Benchmark & DCLM-RW & Nemo-CC \\
\midrule
TriviaQA      & 15.1 & 14.6 \\
HellaSwag     & 12.3 & 12.3 \\
MMLU          & 12.3 & 13.2 \\
WebQs  & 12.1 & 11.3 \\
WinoGrande    & 12.0 & 10.7 \\
PIQA          & 11.3 & 11.5 \\
Lambada       & 6.6  & 5.8 \\
SciQ          & 6.4  & 7.0 \\
ARC-Challenge & 6.2  & 7.0 \\
ARC-Easy      & 5.8  & 6.5 \\
\bottomrule
\end{tabular}
\caption{\textbf{Benchmark contribution to BETR document selection.} Percentage of top-10\% documents attributed to each benchmark when using BETR Target-\coremetric with max aggregation. Each document is attributed to the benchmark whose example ranks it highest. All benchmarks are equally targeted (10\% of total examples each) but contribute differently to top-10\% selection.  \textcolor{tabgray}{DCLM-RW = DCLM-RefinedWeb. Nemo-CC = Nemotron-CC.}}
\label{tab:app_benchmark_contribution}
\end{table}

%% file: figures/appendix/topic_format_distribution.tex
\begin{figure}[!ht]
\centering
\includegraphics[width=\columnwidth]{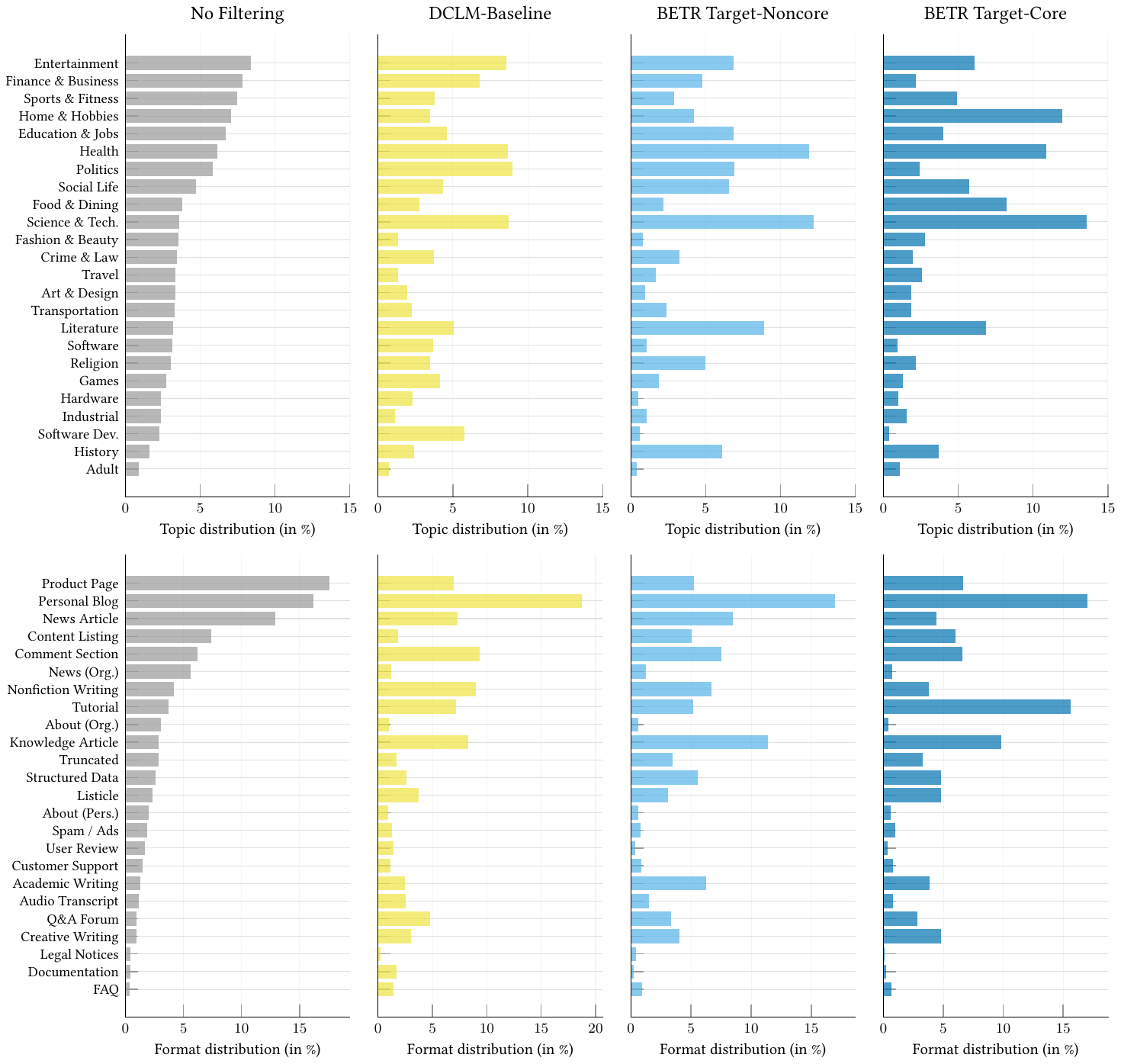}
\caption{\textbf{Topic and format distributions on DCLM-RefinedWeb.} We show the WebOrganizer~\citep{wettig2025organize} topic and format distributions for four datasets: unfiltered DCLM-RefinedWeb, DCLM-Baseline, BETR Target-\extendedmetric, and BETR Target-\coremetric (the latter three filtered to top 10\% of tokens).}
\label{fig:app_topic_format_distribution}
\end{figure}

%% file: tables/appendix/noncore_benchmark_list.tex
\begin{table}[htbp]
\centering
\small
\adjustbox{max width=\textwidth}{%
\begin{tabular}{llccl}
\toprule
Benchmark & Subcategory & Order A & Order B & Source \\
\midrule
AGIEval LSAT AR & Symbolic PS & 1 & 39 & \citet{zhong2023agieval} \\
AGIEval LSAT LR & Reading Comp. & 20 & 5 & \citet{zhong2023agieval} \\
AGIEval LSAT RC & Reading Comp. & 39 & 2 & \citet{zhong2023agieval} \\
AGIEval SAT EN & Reading Comp. & 39 & 39 & \citet{zhong2023agieval} \\
AGIEval SAT Math & Symbolic PS & 39 & 39 & \citet{zhong2023agieval} \\
AQuA & Symbolic PS & 20 & 10 & \citet{ling2017aqua} \\
BBQ & Commonsense R. & 39 & 39 & \citet{parrish2021bbq} \\
BIG-bench Conceptual Combinations & Lang. Und. & 20 & 39 & \citet{srivastava2023beyond} \\
BIG-bench Conlang Translation & Lang. Und. & 20 & 39 & \citet{srivastava2023beyond} \\
BIG-bench CS Algorithms & Symbolic PS & 39 & 5 & \citet{srivastava2023beyond} \\
BIG-bench Dyck Languages & Symbolic PS & 20 & 20 & \citet{srivastava2023beyond} \\
BIG-bench Elementary Math QA & Symbolic PS & 39 & 5 & \citet{srivastava2023beyond} \\
BIG-bench Language Identification & Lang. Und. & 5 & 39 & \citet{srivastava2023beyond} \\
BIG-bench Logical Deduction & Symbolic PS & 10 & 39 & \citet{srivastava2023beyond} \\
BIG-bench Misconceptions & World Knowledge & 10 & 20 & \citet{srivastava2023beyond} \\
BIG-bench Novel Concepts & Commonsense R. & 39 & 39 & \citet{srivastava2023beyond} \\
BIG-bench Operators & Symbolic PS & 39 & 39 & \citet{srivastava2023beyond} \\
BIG-bench Repeat Copy Logic & Symbolic PS & 20 & 10 & \citet{srivastava2023beyond} \\
BIG-bench Strange Stories & Commonsense R. & 39 & 20 & \citet{srivastava2023beyond} \\
BIG-bench Strategy QA & Commonsense R. & 39 & 39 & \citet{srivastava2023beyond} \\
BIG-bench Understanding Fables & Reading Comp. & 10 & 39 & \citet{srivastava2023beyond} \\
BoolQ & Reading Comp. & 20 & 20 & \citet{clark2019boolq} \\
CommonsenseQA & Commonsense R. & 20 & 20 & \citet{talmor2018commonsenseqa} \\
COPA & Commonsense R. & 20 & 10 & \citet{roemmele2011copa} \\
CoQA & Reading Comp. & 39 & 39 & \citet{reddy2019coqa} \\
Enterprise PII Classification & Lang. Und. & 39 & 39 & \citet{patronusai2023enterprisepii} \\
GPQA Diamond & World Knowledge & 39 & 20 & \citet{rein2024gpqa} \\
GPQA Main & World Knowledge & 5 & 10 & \citet{rein2024gpqa} \\
GSM8K & Symbolic PS & 20 & 39 & \citet{cobbe2021gsm8k} \\
Jeopardy & World Knowledge & 10 & 39 & \citet{tunguz2019jeopardy} \\
LogiQA & Symbolic PS & 39 & 39 & \citet{liu2020logiqa} \\
MathQA & Symbolic PS & 39 & 39 & \citet{amini2019mathqa} \\
OpenBookQA & Commonsense R. & 39 & 20 & \citet{mihaylov2018obqa} \\
PubMedQA & Reading Comp. & 39 & 20 & \citet{jin2019pubmedqa} \\
Simple Arithmetic No Spaces & Symbolic PS & 5 & 20 & \citet{mosaicml2023evalgauntlet} \\
Simple Arithmetic With Spaces & Symbolic PS & 39 & 10 & \citet{mosaicml2023evalgauntlet} \\
Social IQA & Commonsense R. & 10 & 39 & \citet{sap2019socialiqa} \\
SQuAD & Reading Comp. & 2 & 1 & \citet{rajpurkar2016squad} \\
SVAMP & Symbolic PS & 39 & 20 & \citet{patel2021svamp} \\
\bottomrule
\end{tabular}%
}
\caption{\textbf{List of the 39 \extendedmetric benchmarks} (from DCLM-Core and DCLM-Extended~\citep{li2024dclm}, excluding overlap with \coremetric).  Order A and Order B indicate each benchmark's position in two random orderings used to test how performance scales with benchmark diversity (1, 2, 5, 10, 20, and 39 benchmarks) in Figure~\ref{fig:evaluation_blind_progression}. \textcolor{tabgray}{Commonsense R. = Commonsense Reasoning, Lang. Und. = Language Understanding, Reading Comp. = Reading Comprehension, Symbolic PS = Symbolic Problem Solving.}}
\label{tab:app_noncore_benchmark_list}
\end{table}

%% file: tables/appendix/loss_scaling_law_parameters.tex
\begin{table}[tpb]
\centering
\begin{adjustbox}{width=\textwidth}
{%
  \setlength{\tabcolsep}{2pt}      %
\renewcommand{\arraystretch}{1.0}  %
    \begin{tabular}{
        l                   %
        @{\hspace{8pt}}     %
        *{5}{c}             %
        @{\hspace{5pt}}     %
        G                   %
    }
    \toprule
    Dataset & a & b & e & $\alpha$ & $\beta$ & MAE \\
    \midrule
    \multicolumn{7}{l}{\textbf{Data pool: Nemotron-CC, Filter: No Filter}} \\ \midrule
    ARC-Easy & 4.092 (3.370, 5.743) & 6.906 (5.607, 9.345) & -0.9782 (-1.248, -0.6162) & 0.2454 (0.1992, 0.3478) & 0.3628 (0.2961, 0.4821) & 0.018  \\
    ARC-Challenge & 3.853 (2.586, 5.435) & 7.026 (4.892, 8.464) & -0.7305 (-1.246, -0.5034) & 0.2407 (0.1572, 0.3377) & 0.3764 (0.2668, 0.4456) & 0.016  \\
    HellaSwag & 5.001 (4.291, 5.936) & 6.544 (6.395, 6.739) & -0.5320 (-0.5638, -0.4977) & 0.3519 (0.3092, 0.4071) & 0.3864 (0.3794, 0.3965) & 0.005  \\
    Lambada & 5.669 (3.632, 6.944) & 7.834 (5.083, 9.892) & -1.787 (-12.92, -1.340) & 0.3399 (0.2196, 0.4161) & 0.4089 (0.2746, 0.5129) & 0.012  \\
    MMLU & 5.039 (2.832, 6.256) & 5.620 (4.617, 8.338) & -0.5209 (-0.9907, -0.3759) & 0.3157 (0.1740, 0.3851) & 0.3034 (0.2514, 0.4412) & 0.015  \\
    PIQA & 6.092 (5.224, 6.751) & 6.314 (6.044, 7.174) & -0.3418 (-0.3790, -0.3068) & 0.3999 (0.3489, 0.4389) & 0.3624 (0.3477, 0.4064) & 0.006  \\
    SciQ & 3.550 (2.895, 4.413) & 5.473 (3.696, 8.820) & -106.1 (-106.1, -106.1) & 0.1940 (0.1538, 0.2474) & 0.2707 (0.1831, 0.4357) & 0.043  \\
    TriviaQA & 3.427 (1.969, 5.385) & 6.520 (4.686, 7.509) & 0.1640 (-2.863, 0.3539) & 0.1962 (0.07290, 0.3200) & 0.3339 (0.2390, 0.3828) & 0.023  \\
    WebQs & 5.922 (4.587, 7.566) & 8.420 (5.951, 9.234) & -0.5823 (-0.7948, -0.4336) & 0.3526 (0.2759, 0.4494) & 0.4299 (0.3079, 0.4704) & 0.018  \\
    WinoGrande & 3.449 (3.448, 3.449) & 4.840 (4.839, 4.840) & -1.291 (-1.300, -1.278) & 0.3424 (0.3384, 0.3478) & 0.3738 (0.3713, 0.3782) & 0.002  \\
    Val Loss & 5.450 (4.608, 6.971) & 6.676 (6.174, 7.641) & 0.5218 (0.4453, 0.6148) & 0.3037 (0.2518, 0.3945) & 0.3282 (0.3029, 0.3780) & 0.022  \\
    \midrule
    \multicolumn{7}{l}{\textbf{Data pool: Nemotron-CC, Filter: BETR Target-\coremetric Top 3\%}} \\ \midrule
    ARC-Easy & 4.480 (3.464, 5.976) & 8.105 (5.881, 8.658) & -0.8082 (-1.055, -0.6930) & 0.2954 (0.2310, 0.3834) & 0.4472 (0.3341, 0.4739) & 0.012  \\
    ARC-Challenge & 4.651 (3.681, 5.474) & 7.431 (6.411, 7.951) & -0.6484 (-0.7877, -0.5992) & 0.3095 (0.2503, 0.3578) & 0.4150 (0.3631, 0.4409) & 0.009  \\
    HellaSwag & 5.111 (4.374, 5.955) & 6.630 (6.432, 6.764) & -0.5466 (-0.5779, -0.5238) & 0.3614 (0.3172, 0.4105) & 0.3943 (0.3842, 0.4013) & 0.004  \\
    Lambada & 6.923 (5.576, 7.801) & 10.45 (8.896, 11.97) & -1.650 (-1.832, -1.482) & 0.4176 (0.3431, 0.4697) & 0.5401 (0.4650, 0.6120) & 0.011  \\
    MMLU & 4.359 (3.525, 6.152) & 6.243 (5.439, 8.014) & -0.5874 (-0.7090, -0.4056) & 0.2830 (0.2308, 0.3919) & 0.3424 (0.3021, 0.4326) & 0.012  \\
    PIQA & 5.565 (5.565, 6.239) & 6.913 (6.781, 6.914) & -0.3924 (-0.4109, -0.3776) & 0.3858 (0.3821, 0.4247) & 0.4083 (0.4018, 0.4102) & 0.006  \\
    SciQ & 6.735 (3.341, 10.20) & 15.48 (9.509, 19.85) & -0.6378 (-1.523, -0.5290) & 0.3992 (0.1906, 0.5964) & 0.7822 (0.4912, 0.9972) & 0.045  \\
    TriviaQA & 3.732 (2.215, 5.297) & 7.449 (5.828, 8.053) & 0.2570 (-0.4100, 0.3792) & 0.2174 (0.1025, 0.3161) & 0.3809 (0.2996, 0.4103) & 0.021  \\
    WebQs & 7.586 (5.926, 8.412) & 8.494 (7.333, 9.468) & -0.4430 (-0.5773, -0.4059) & 0.4591 (0.3637, 0.5053) & 0.4383 (0.3803, 0.4862) & 0.013  \\
    WinoGrande & 3.301 (3.301, 3.302) & 4.951 (4.951, 4.951) & -1.296 (-1.306, -1.284) & 0.3379 (0.3315, 0.3420) & 0.3810 (0.3774, 0.3849) & 0.002  \\
    Val Loss & 5.291 (4.456, 7.190) & 6.294 (5.409, 7.744) & 0.2539 (0.1408, 0.3988) & 0.2980 (0.2462, 0.4099) & 0.3127 (0.2662, 0.3873) & 0.022  \\
    \midrule
    \multicolumn{7}{l}{\textbf{Data pool: Nemotron-CC, Filter: BETR Target-\coremetric Top 10\%}} \\ \midrule
    ARC-Easy & 5.739 (4.036, 7.209) & 9.396 (8.198, 10.96) & -0.6646 (-0.8457, -0.5862) & 0.3626 (0.2611, 0.4485) & 0.5026 (0.4441, 0.5757) & 0.013  \\
    ARC-Challenge & 4.864 (3.799, 5.869) & 9.246 (8.367, 9.957) & -0.6004 (-0.7120, -0.5320) & 0.3151 (0.2497, 0.3758) & 0.4953 (0.4523, 0.5308) & 0.010  \\
    HellaSwag & 5.168 (4.615, 5.963) & 6.428 (6.260, 6.680) & -0.5568 (-0.5757, -0.5278) & 0.3656 (0.3337, 0.4128) & 0.3838 (0.3758, 0.3979) & 0.004  \\
    Lambada & 6.778 (5.227, 7.403) & 9.920 (7.617, 10.92) & -1.620 (-2.073, -1.502) & 0.4110 (0.3230, 0.4468) & 0.5171 (0.4049, 0.5663) & 0.008  \\
    MMLU & 5.188 (4.390, 6.324) & 7.128 (5.988, 8.605) & -0.5057 (-0.6094, -0.4391) & 0.3295 (0.2812, 0.3948) & 0.3847 (0.3263, 0.4587) & 0.011  \\
    PIQA & 6.086 (4.986, 6.086) & 6.907 (6.754, 6.907) & -0.3808 (-0.4032, -0.3677) & 0.4104 (0.3488, 0.4149) & 0.4019 (0.3942, 0.4065) & 0.006  \\
    SciQ & 5.417 (3.766, 6.138) & 12.29 (9.624, 15.88) & -0.8275 (-1.434, -0.7324) & 0.3180 (0.2140, 0.3593) & 0.6231 (0.4914, 0.7901) & 0.030  \\
    TriviaQA & 3.983 (2.306, 5.237) & 5.637 (4.719, 7.557) & 0.1363 (-0.6256, 0.3053) & 0.2305 (0.1093, 0.3098) & 0.2873 (0.2378, 0.3837) & 0.023  \\
    WebQs & 7.187 (6.225, 7.731) & 9.279 (7.899, 9.725) & -0.4878 (-0.6013, -0.4695) & 0.4332 (0.3758, 0.4633) & 0.4754 (0.4060, 0.4961) & 0.012  \\
    WinoGrande & 3.568 (3.567, 3.568) & 4.647 (4.647, 4.647) & -1.299 (-1.307, -1.287) & 0.3541 (0.3515, 0.3610) & 0.3681 (0.3651, 0.3720) & 0.002  \\
    Val Loss & 5.510 (4.608, 7.076) & 6.680 (6.067, 7.709) & 0.4046 (0.3156, 0.5033) & 0.3093 (0.2536, 0.4027) & 0.3300 (0.2981, 0.3835) & 0.021  \\
    \midrule
    \multicolumn{7}{l}{\textbf{Data pool: Nemotron-CC, Filter: BETR Target-\coremetric Top 30\%}} \\ \midrule
    ARC-Easy & 6.079 (4.420, 6.847) & 8.040 (6.119, 9.260) & -0.6383 (-0.8433, -0.5876) & 0.3800 (0.2805, 0.4242) & 0.4258 (0.3319, 0.4834) & 0.013  \\
    ARC-Challenge & 4.437 (3.762, 5.761) & 7.637 (6.121, 8.494) & -0.6142 (-0.7315, -0.5305) & 0.2883 (0.2471, 0.3668) & 0.4129 (0.3369, 0.4543) & 0.011  \\
    HellaSwag & 5.057 (4.507, 5.779) & 6.690 (6.510, 6.808) & -0.5476 (-0.5713, -0.5252) & 0.3577 (0.3252, 0.3999) & 0.3968 (0.3879, 0.4030) & 0.004  \\
    Lambada & 6.736 (4.803, 7.601) & 9.343 (8.454, 11.64) & -1.558 (-2.021, -1.357) & 0.4061 (0.2953, 0.4571) & 0.4882 (0.4441, 0.6004) & 0.012  \\
    MMLU & 4.501 (4.033, 5.774) & 7.398 (6.430, 8.098) & -0.5326 (-0.5753, -0.4419) & 0.2874 (0.2582, 0.3623) & 0.3959 (0.3494, 0.4307) & 0.011  \\
    PIQA & 5.799 (4.905, 5.800) & 6.843 (6.587, 6.957) & -0.3681 (-0.4037, -0.3592) & 0.3904 (0.3380, 0.3935) & 0.3944 (0.3809, 0.4002) & 0.005  \\
    SciQ & 3.811 (3.170, 6.230) & 7.890 (5.179, 11.24) & -2.408 (-14.80, -0.8117) & 0.2092 (0.1697, 0.3603) & 0.3958 (0.2607, 0.5617) & 0.032  \\
    TriviaQA & 3.257 (2.282, 5.783) & 5.930 (4.151, 8.014) & 0.007347 (-1.014, 0.3278) & 0.1819 (0.1058, 0.3435) & 0.3026 (0.2081, 0.4092) & 0.025  \\
    WebQs & 6.101 (4.988, 7.933) & 7.988 (6.765, 9.796) & -0.6492 (-0.7833, -0.4710) & 0.3666 (0.3012, 0.4707) & 0.4063 (0.3482, 0.4963) & 0.017  \\
    WinoGrande & 2.922 (2.922, 2.922) & 6.419 (6.419, 6.419) & -1.299 (-1.304, -1.287) & 0.3148 (0.3133, 0.3195) & 0.4550 (0.4519, 0.4597) & 0.001  \\
    Val Loss & 5.350 (4.554, 6.861) & 6.828 (6.298, 7.582) & 0.5015 (0.4257, 0.5969) & 0.2975 (0.2482, 0.3881) & 0.3362 (0.3103, 0.3757) & 0.022  \\
    \bottomrule
    \end{tabular}
}%
\end{adjustbox}
\caption{\textbf{Loss scaling law parameters} for BETR Target-\coremetric and no filtering baseline on the Nemotron-CC data pool. Each column contains the estimated value and the 95\% confidence range (lower bound, upper bound). The last column shows the mean average error (MAE) for the fit evaluated on the training date.}
\label{tab:app_loss_scaling_law_params}
\end{table}

%% file: tables/appendix/acc_scaling_law_parameters.tex
\begin{table}[tpb]
\centering
\begin{adjustbox}{width=\textwidth}
{%
  \setlength{\tabcolsep}{2pt}      %
\renewcommand{\arraystretch}{1.0}  %
    \begin{tabular}{
        l                   %
        @{\hspace{8pt}}     %
        *{4}{c}             %
        @{\hspace{5pt}}     %
        G                   %
    }
    \toprule
    Dataset & $c_1$ & $c_2$ & $k$ & $L_0$ & MAE  \\
    \midrule
    \multicolumn{6}{l}{\textbf{Data pool: Nemotron-CC, Filter: No Filter}} \\ \midrule
    ARC-Easy& 0.6722 (0.4884, 1.0000)& 0.3558 (0.2785, 0.4079)& -3.749 (-5.239, -2.582)& 0.8376 (0.6798, 0.9131) & 0.0096  \\
    ARC-Challenge& 0.8419 (0.5784, 1.0000)& 0.1800 (0.1646, 0.1920)& -5.792 (-6.887, -5.026)& 0.6264 (0.5699, 0.7293) & 0.0075  \\
    HellaSwag& 0.7488 (0.4336, 0.7563)& 0.2520 (0.2449, 0.2756)& -10.01 (-14.43, -9.628)& 0.6357 (0.6342, 0.7124) & 0.0024  \\
    Lambada& 1.0000 (0.7478, 1.0000)& 0.1774 (0.1505, 0.1954)& -3.977 (-4.774, -3.690)& 0.3516 (0.3386, 0.4691) & 0.0061  \\
    MMLU& 0.9011 (0.3020, 1.0000)& 0.2305 (0.2268, 0.2415)& -4.207 (-6.040, -3.954)& 0.4197 (0.3755, 0.7764) & 0.0022  \\
    PIQA& 0.4396 (0.2719, 1.0000)& 0.5683 (0.5206, 0.6025)& -4.951 (-7.662, -3.036)& 0.8075 (0.4770, 0.9077) & 0.0038  \\
    SciQ& 1.0000 (0.2714, 1.0000)& 0.02759 (3.7659e-07, 0.6920)& -1.634 (-3.935, -1.472)& 2.113 (1.262, 2.162) & 0.0081  \\
    TriviaQA& 0.9998 (0.4204, 1.0000)& 3.4328e-10 (4.9324e-26, 0.01807)& -4.385 (-7.159, -4.275)& 1.492 (1.482, 1.789) & 0.0053  \\
    WebQs& 1.0000 (0.3478, 1.0000)& 0.02107 (3.1107e-18, 0.02739)& -5.668 (-6.308, -4.309)& 0.5104 (0.4521, 0.7950) & 0.0071  \\
    WinoGrande& 0.5039 (0.1537, 0.5154)& 0.4950 (0.4852, 0.5137)& -57.76 (-153.8, -49.01)& 0.2862 (0.2833, 0.3135) & 0.0087  \\
    \midrule
    \multicolumn{6}{l}{\textbf{Data pool: Nemotron-CC, Filter: BETR Target-\coremetric Top 3\%}} \\ \midrule
    ARC-Easy& 0.8852 (0.4598, 1.0000)& 0.1498 (0.03160, 0.4605)& -3.278 (-6.289, -2.741)& 0.9709 (0.7953, 1.051) & 0.0083  \\
    ARC-Challenge& 0.8572 (0.4645, 1.0000)& 0.1709 (0.1401, 0.2321)& -5.470 (-9.441, -4.811)& 0.6166 (0.5647, 0.7427) & 0.0095  \\
    HellaSwag& 0.7556 (0.4923, 0.7637)& 0.2456 (0.2382, 0.2758)& -9.501 (-13.19, -9.210)& 0.6385 (0.6375, 0.6984) & 0.0022  \\
    Lambada& 1.0000 (0.7284, 1.0000)& 0.2017 (0.1620, 0.2287)& -3.924 (-4.896, -3.614)& 0.3181 (0.2999, 0.4411) & 0.0062  \\
    MMLU& 0.8737 (0.3396, 0.9120)& 0.2297 (0.2245, 0.2473)& -4.363 (-6.527, -4.132)& 0.4603 (0.4360, 0.7552) & 0.0020  \\
    PIQA& 0.4379 (0.2018, 1.0000)& 0.5733 (0.4199, 0.6569)& -4.576 (-10.40, -2.461)& 0.7951 (0.6157, 0.8970) & 0.0038  \\
    SciQ& 0.2032 (0.1559, 1.0000)& 0.8024 (0.4640, 0.8325)& -4.286 (-6.197, -0.8398)& 0.9061 (0.09884, 1.101) & 0.0077  \\
    TriviaQA& 1.0000 (0.4673, 1.0000)& 3.9441e-10 (1.2915e-26, 0.02202)& -3.800 (-6.027, -3.684)& 1.557 (1.544, 1.856) & 0.0074  \\
    WebQs& 1.0000 (0.1760, 1.0000)& 0.04284 (2.1853e-17, 0.05154)& -5.305 (-9.153, -4.109)& 0.4699 (0.4552, 0.9828) & 0.0111  \\
    WinoGrande& 0.5008 (0.2141, 0.5132)& 0.4980 (0.4868, 0.5146)& -63.36 (-118.8, -56.24)& 0.2870 (0.2859, 0.3067) & 0.0080  \\
    \midrule
    \multicolumn{6}{l}{\textbf{Data pool: Nemotron-CC, Filter: BETR Target-\coremetric Top 10\%}} \\ \midrule
    ARC-Easy& 0.5973 (0.3603, 1.0000)& 0.4175 (0.2657, 0.5268)& -4.606 (-8.508, -2.683)& 0.7961 (0.6818, 0.8509) & 0.0102  \\
    ARC-Challenge& 0.8554 (0.4957, 1.0000)& 0.1719 (0.1474, 0.2133)& -5.509 (-8.254, -4.763)& 0.6191 (0.5671, 0.7461) & 0.0077  \\
    HellaSwag& 0.7686 (0.4782, 0.7814)& 0.2329 (0.2209, 0.2707)& -9.204 (-13.58, -8.814)& 0.6378 (0.6363, 0.7027) & 0.0026  \\
    Lambada& 1.0000 (0.7462, 1.0000)& 0.2078 (0.1751, 0.2381)& -4.002 (-4.967, -3.747)& 0.3159 (0.3027, 0.4204) & 0.0060  \\
    MMLU& 0.8150 (0.2729, 1.0000)& 0.2407 (0.2325, 0.2492)& -5.136 (-7.202, -4.511)& 0.5086 (0.4367, 0.7988) & 0.0029  \\
    PIQA& 0.4325 (0.2592, 1.0000)& 0.5786 (0.3826, 0.6342)& -4.622 (-7.839, -2.030)& 0.7883 (0.4165, 0.8748) & 0.0041  \\
    SciQ& 0.2774 (0.1862, 1.0000)& 0.7368 (0.1267, 0.8164)& -3.241 (-4.881, -0.8278)& 1.036 (0.5095, 1.965) & 0.0068  \\
    TriviaQA& 0.9983 (0.5154, 1.0000)& 1.0469e-23 (8.9296e-29, 0.01735)& -4.015 (-6.040, -3.908)& 1.525 (1.515, 1.784) & 0.0080  \\
    WebQs& 1.0000 (0.2636, 1.0000)& 0.03153 (7.7517e-16, 0.04030)& -5.364 (-7.112, -4.443)& 0.4882 (0.4753, 0.8897) & 0.0100  \\
    WinoGrande& 0.4989 (0.2339, 0.5116)& 0.4995 (0.4897, 0.5147)& -67.61 (-122.1, -59.93)& 0.2864 (0.2849, 0.3032) & 0.0087  \\
    \midrule
    \multicolumn{6}{l}{\textbf{Data pool: Nemotron-CC, Filter: BETR Target-\coremetric Top 30\%}} \\ \midrule
    ARC-Easy& 0.6267 (0.5370, 1.0000)& 0.3909 (0.2106, 0.4662)& -4.516 (-5.633, -2.713)& 0.8159 (0.6307, 0.8790) & 0.0103  \\
    ARC-Challenge& 0.8254 (0.6361, 1.0000)& 0.1872 (0.1673, 0.2067)& -6.502 (-7.797, -5.560)& 0.6432 (0.5874, 0.7043) & 0.0086  \\
    HellaSwag& 0.7639 (0.4908, 0.7753)& 0.2373 (0.2267, 0.2677)& -9.501 (-13.31, -9.114)& 0.6376 (0.6363, 0.6998) & 0.0024  \\
    Lambada& 1.0000 (0.7984, 1.0000)& 0.1868 (0.1665, 0.2015)& -4.063 (-4.723, -3.857)& 0.3451 (0.3337, 0.4367) & 0.0066  \\
    MMLU& 0.7986 (0.4090, 1.0000)& 0.2416 (0.2350, 0.2469)& -5.488 (-6.374, -4.907)& 0.5351 (0.4602, 0.7061) & 0.0022  \\
    PIQA& 0.4739 (0.4057, 1.0000)& 0.5406 (0.2289, 0.5993)& -4.204 (-5.392, -1.901)& 0.8265 (0.4495, 1.012) & 0.0043  \\
    SciQ& 0.2458 (0.2036, 1.0000)& 0.7637 (0.4928, 0.7940)& -3.410 (-4.347, -0.8965)& 0.9745 (0.09342, 1.157) & 0.0075  \\
    TriviaQA& 0.9966 (0.4876, 1.0000)& 1.9465e-23 (2.3055e-29, 0.01879)& -4.214 (-6.792, -4.084)& 1.500 (1.488, 1.767) & 0.0084  \\
    WebQs& 1.0000 (0.2758, 1.0000)& 0.02252 (0.006499, 0.03034)& -5.567 (-8.836, -4.952)& 0.5043 (0.4865, 0.8482) & 0.0078  \\
    WinoGrande& 0.5049 (0.2222, 0.5239)& 0.4945 (0.4828, 0.5069)& -65.29 (-109.2, -57.86)& 0.2865 (0.2852, 0.3054) & 0.0090  \\
    \bottomrule
    \end{tabular}
}%
\end{adjustbox}
\caption{\textbf{Accuracy scaling law parameters} for BETR Target-\coremetric and no filtering baseline on the Nemotron-CC data pool. Each column contains the estimated value and the 95\% confidence range (lower bound, upper bound). The last column shows the mean average error (MAE) for the fit evaluated on the training data.
}
\label{tab:app_acc_scaling_law_params}
\end{table}